\documentclass[lettersize, journal]{IEEEtran}
\usepackage{cite}
\usepackage{amsmath,amssymb,amsfonts}
\usepackage[ruled,vlined,noend]{algorithm2e}
\usepackage{graphicx}
\usepackage{textcomp}
\usepackage{tabularx}

\usepackage[]{units}
\usepackage{bm}
\usepackage{amssymb}
\usepackage{url}


\usepackage{caption}
\usepackage{subcaption}
\usepackage{algpseudocode}
\usepackage{paralist}
\usepackage{mathtools}

\usepackage{multirow}
\usepackage{hyperref}
\usepackage{amsmath} 
\usepackage{amssymb}  
\usepackage{mathtools}
\DeclareMathOperator*{\argmin}{arg\,min}
\usepackage[acronym,nomain,nonumberlist]{glossaries}

\def\BibTeX{{\rm B\kern-.05em{\sc i\kern-.025em b}\kern-.08em
    T\kern-.1667em\lower.7ex\hbox{E}\kern-.125emX}}
    
\begin{document}
\markboth{IEEE Transactions on Robotics, VOL. XX, NO. XX, XXXX 2023}
{Lindqvist \MakeLowercase{\textit{et al.}}: Tree-based Exploration}

\title{A Tree-based Next-best-trajectory Method for 3D UAV Exploration}
\author{Björn Lindqvist$^{*1}$, Akash Patel$^{*1}$, Kalle Löfgren$^2$, and George Nikolakopoulos$^1$
\thanks{Manuscript received: 13-05-2023. Revision received: 29-02-2024. Accepted: 27-05-2024}
\thanks{This work has been partially funded by the European Unions Horizon 2020 Research and Innovation Programme under the Grant Agreement No. 869379 illuMINEation and No.101003591 NEXGEN-SIMS, and in part by the Swedish Department of Energy and LKAB through the Sustainable Underground Mining (SUM) Academy Programme project "Autonomous Drones for Underground Mining Operations".} 
\thanks{$^1$Bj\"orn Lindqvist, Akash Patel, and George Nikolakopoulos are with the Robotics and Artificial Intelligence Group, Department of Computer, Electrical and Space Engineering, Lule\r{a} University of Technology, Lule\r{a} SE-97187, Sweden.}
\thanks{$^2$Kalle Löfgren did his Masters Thesis in Computer Science Engineering at the Department of Computer, Electrical and Space Engineering, Lule\r{a} University of Technology, in collaboration with the Robotics and AI Group.}
\thanks{Emails of authors in order of appearance: \texttt{bjolin@ltu.se , akapat@ltu.se, kallelofgren94@gmail.com, geonik@ltu.se.}}
\thanks{*Authors contributed Equally}}

\maketitle

\begin{abstract}
This work presents a fully integrated tree-based combined exploration-planning algorithm: Exploration-RRT (ERRT). The algorithm is focused on providing real-time solutions for local exploration in a fully unknown and unstructured environment while directly incorporating exploratory behavior, robot-safe path planning, and robot actuation into the central problem. ERRT provides a complete sampling and tree-based solution for evaluating "where to go next" by considering a trade-off between maximizing information gain, and minimizing the distances travelled and the robot actuation along the path. The complete scheme is evaluated in extensive simulations, comparisons, as well as real-world field experiments in constrained and narrow subterranean and GPS-denied environments. The framework is fully ROS-integrated, straight-forward to use, and we open-source it at \url{https://github.com/LTU-RAI/ExplorationRRT}.

\end{abstract}

\begin{IEEEkeywords}
Tree-based Exploration, RRT, Subterranean Exploration, Field Robotics, Unmanned Aerial Vehicles
\end{IEEEkeywords}

\section{Introduction}\label{sec:intro}

\subsection{Background}
\IEEEPARstart{T}{he} exploration of unknown environments has long been one of the central problem in robotics from an application perspective in scenarios such as rescue robotics~\cite{dang2020autonomous,lindqvist2022compra}, or for mapping unknown cave systems\cite{petravcek2021large} or urban indoor environments\cite{kratky2021autonomous}. While the ability to operate in known environments has its own set of related challenges, the ability for an autonomous robot to operate in completely unknown areas sets the bar very high when it comes to the onboard decision-making process and autonomy framework. In such scenarios the robot must decide the optimal exploration route based only on its current knowledge of its surroundings, linked directly to onboard perception systems, and update it accordingly as more areas as explored. In other words the fundamental question of "where to go next" has to be answered. During the DARPA Subterranean Challenge \cite{subt} the answer to that question was significantly pushed forward~\cite{agha2021nebula, rouvcek2019darpa, tranzatto2022cerberus}, as teams deployed fleets of robots to explore vast unknown subterranean environments in the search-and-rescue context. 

As a general principle, robot exploration algorithms should guide the robot towards unknown areas often denoted as \textit{frontier points}\cite{yamauchi1997frontier}. Frontier exploration\cite{lluvia2021active} consists of classifying certain areas in a map as frontiers (the boundary between known and unknown space), selecting the best one on some predefined criteria, and then querying a path planner to plan a path from the robot current position to the best frontier. In many of these methods the problem of exploration and the problem of path planning are solved separately which allows both complex frontier selection and detailed path planning. The method in \cite{niroui2019deep} formulated a Neural Network that could classify the frontier point with the highest probability of information gain and use a $\mathrm{A}^*$ planner~\cite{duchoe2014path} to generate the path to that frontier. The work in \cite{patel2023ref} selects frontiers not from a perspective of maximizing information gain, but from a perspective of fuel efficiency and a continued forward exploration, while a risk-aware grid-search algorithm \cite{karlsson2023d+} plans the path. The work in \cite{zhou2021fuel} generates frontiers and plans a global route as a travelling salesman problem, while local kinodynamic paths are planned via B-Spline planning~\cite{zhou2019robust}. The work in \cite{liu2022efficient} also include concepts related to semantic environment representations in classifying frontier points for a more "human-like" exploration process. 

In conclusion, splitting robot exploration into the two separate problems of \textit{explore} and then \textit{plan} allows a wide range of solutions towards optimizing the resulting exploration route for a variety of conditions. But a common drawback of such methods is that how to get to the frontier point is not considered in the frontier evaluation process, which in many situations effects which frontier-point should be explored next. 
\subsection{The combined exploration-planning Problem}\label{sec:combined}
A separate line of research for robot exploration or reconstruction missions, often denoted as as Next-Best-View  methods~\cite{pito1999solution}\cite{low2006efficient}, operate (in generalising terms) by instead of directly selecting an optimal frontier point, they sample a large number of possible poses in the free space around the robot and based on a model of an onboard sensor (camera, LiDAR) select the next best pose (or view) to go to for maximizing the explored volume (maximizing the \textit{information gain}). An outcome of this type of framework were sampling-based methods where, through Rapidly-Exploring Random Trees\cite{kuffner2000rrt} or other graph-based or sampling-based methods, not only a large number of poses but a large number of paths to various poses could be evaluated for maximizing information gain while minimizing the length of the path to get there~\cite{bircher2016receding}. The result was a sampling-based solution to the combined exploration-planning problem.

The ability to solve the exploration-planning combined problem comes at a concession to computational complexity, and how complex the evaluation process of possible paths can reasonably be. Tree- or graph-solutions are efficient at doing this as we can step through nodes in the generated tree/graph and evaluate each branch for parameters such as path length and sensor-based information gain. As such, some next-best-view methods have previously utilized Tree-based solutions~\cite{vasquez2018tree}, and it is the cornerstone of the proposed ERRT framework as well. The work in \cite{bircher2016receding} samples poses (position and heading) states and integrated a camera sensor model for evaluating the predicted explored volume, and solves it as a receding horizon problem showing good results. Similarly, the work in \cite{schmid2020efficient} targets the coverage problem with a camera imprint using RRT integrated with information gain, and with the goal of global coverage. The significant work in \cite{dharmadhikari2020motion} tackled the problem of agile navigation by sampling actuation (acceleration) directly, while again evaluating the resulting graph for the path total length and information gain, while also generating robot-safe paths. This type of solution has seen success in field applications as well~\cite{dang2019graph, dang2020graph} as local and global exploration modules, here denoted as Graph-based solutions as opposed to the RRT.
It should be noted that these concepts of sampling-based \textit{information-gathering planners} are not only used for general robot exploration of an unknown space, but can also be applied to use-cases such as mapping signal strength~\cite{hollinger2014sampling}, or mapping the intensity of a magnetic field~\cite{viseras2017online}. Those cited works are perhaps the most similar to ERRT in their problem formulation set-up of finding the Next-Best-Trajectory but are applied to very different use-cases where information-gain calculations can be performed on the whole tree or graph with less computational issues.

\subsection{Exploration-RRT}
Our previous work in \cite{lindqvist2021exploration}, that presented the initial formulation of ERRT and is the baseline for this work as well, solves the problem in a slightly different way. As opposed to evaluating each branch or edge node in a graph for information gain and other criteria, which can be very computationally heavy, ERRT samples specific goal positions under certain criteria (e.g. information gain greater than zero) and only evaluates the best RRT-branches that eventually lead to those goals to reduce the computational effort and enabling more complex modules to be added on top (as the total number of paths to be evaluated is pre-defined). In ERRT, those modules are 1) path optimization through iterative environment collision checks to shorten the sampled paths, and 2) solving the predicted robot model-based actuation along the paths as a NMPC (nonlinear model predictive control) problem, from which we both generate dynamics-based paths and have the ability to include the robot actuation along the path in the path evaluation process, and 3) we evaluate the information gain without averaging and \textit{along} the trajectory, not only at the edge of the graph or leaf of the tree. This results in a better approximation of the predicted information gain for following a candidate trajectory. These three modules would be too computationally expensive for a real-time application if run for the whole tree.

 While the over-time exploration process is often described by a general reduction of map uncertainty, the momentary "where to go next" - the next-best-trajectory problem - can be described by:
 \begin{align}\label{eq:realproblem}
    \operatorname*{Minimize}_{
        \mathrm{trajectory}\hspace{0.1cm}\bm{\chi}
    } \, C_d(\bm{\chi}) + C_r(\bm{\chi}) + C_u(\bm{u}(\bm{\chi})) - C_i(\nu(\bm{\chi})) \\
& \hspace{-1.7cm}\text{subj. to:}\hspace{0.1cm}\bm{\chi} \in V_\mathrm{free} \notag \\
& \hspace{-0.5cm}\nu(\bm{\chi}) > 0 \notag
\end{align}
The formulation in \eqref{eq:realproblem} describes the problem we would like to solve for combined exploration-planning. Here, $\bm{\chi}$ represents the state-trajectory to be solved for, while $C_d(\bm{\chi})$ is a cost associated with the path length, $C_r(\bm{\chi})$ a cost associated with the risk or \textit{traversability} of $\bm{\chi}$, $C_u(\bm{u}(\bm{\chi}))$ is the cost related to the model-based actuation along $\bm{\chi}$, and $-C_i(\nu(\bm{\chi}))$ is the revenue or negative costs from the sensor-based information gain $\nu$ along $\bm{\chi}$ (e.g. not just at the end-pose but along the full trajectory). Here, $V_{\mathrm{free}}$ denotes the free space as a subset of $V_\mathrm{map}$ describing an occupancy grid~\cite{hornung2013octomap} of the space around the robot divided into free and occupied space. We note that a linear combination of the cost terms are used in \eqref{eq:realproblem}, as that is how ERRT is implemented for the ease of tuning the relative emphasis on each separate cost, but that is not a requirement and any functional combination of terms could be used without a loss of generality (ex. information gain per distance).
At the moment, there doesn't exist any optimization methods for solving such a minimization problem, and as such we are forced to look at sampling-based methods instead. In ERRT the exploration-planning problem in \eqref{eq:realproblem} is reformulated, under the assumption that if many sampled solutions are analyzed it will approximate \eqref{eq:realproblem}, as:

\begin{align}\label{eq:approx}
   \bm{\chi}^* = \argmin_{
        \hspace{0.1cm}(\bm{\chi}_j)_j
    } \,  (C_d(\bm{\chi}_j) + C_u(\bm{u}(\bm{\chi}_j)) - C_i(\nu(\bm{\chi}_j))_j),\\  j = 1, \ldots, n_\mathrm{traj}\notag \\
    \hspace{-1cm}\text{where}~ \bm{\chi}_j \in V_\mathrm{safe}\notag \\
&   \hspace{-1.5cm} \nu(\bm{\chi}_j) > 0 \notag
\end{align}
where $n_\mathrm{traj}$ denotes the number of sampled candidate trajectories that satisfy $\nu(\bm{\chi}) > 0$. Based on the formulation of the problem in \eqref{eq:realproblem} and a high enough number of sampled trajectories $n_\mathrm{traj}$, we denote the output as the approximate optimal next-best-trajectory $\bm{\chi}^*$.
For an UAV that can move freely in 3D, the risk-cost $C_r(\bm{\chi})$ can be simplified into the statement that $\bm{\chi}$ should be in $V_\mathrm{safe}$ where $V_\mathrm{safe} \subset V_\mathrm{free}$ such that for all safe and free voxels $\{F_s\}$ in $V_\mathrm{safe}$:
\begin{subequations}
\begin{align}
    \mid\mid \{F_s\}_i - \{O\}_\mathrm{closest}\mid \mid > r_\mathrm{robot} \\
    \mid\mid \{F_s\}_i - \{U\}_\mathrm{closest}\mid \mid > r_\mathrm{robot}
\end{align}
\end{subequations}
with $\{O\}_\mathrm{closest}$ and $\{U\}_\mathrm{closest}$ denoting the position of the closest occupied or unknown voxel to $\{F_s\}_i$ and $r_\mathrm{robot}$ denotes the size-radius of the robot. It should be noted that not all robot-safe planners adopt this cost-to-constraint re-formulation of $C_r(\bm{\chi})$, and that methods exist that add obstacle costs or risk cost heuristics in the planning problem~\cite{musil2022spheremap, karlsson2023d+, cover2013sparse}. The benefit being that the risk cost can promote the robot to keep a large safety distance but still allows it to move close to an obstacle if absolutely necessary to move through narrow areas. The benefit of the constraint-type statement $\bm{\chi}_j \in V_\mathrm{safe}$ is that the UAV is allowed to explore freely as long as the safety constraint is met without being penalized in the path selection by moving close to obstacles/walls/etc.. As such we are leveraging the UAV mobility in the exploration process, but, perhaps, generating higher-risk paths especially in the presence of uncertainties.
In practise, the safety distance constraint method must also compensate for such uncertainty in localization and control and for some noise in the occupancy mapping process, and should be increased above the real radius of the robot accordingly. Such an increase does lead to the risk of not being able to pass through a constrained entrance.

The sampling minimization in \eqref{eq:approx} describes the end result of the ERRT algorithm, while the Section \ref{sec:implementation} will describe how this is implemented.
%
\subsection{Contributions}
This work presents a continued development and realisation of the ERRT~\cite{lindqvist2021exploration} framework, as well as significant evaluation through realistic simulations, experiments and comparisons. The ERRT framework offers differences and contributions to the state-of-the-art exploration-planning algorithms as follows: 
\begin{itemize}
    \item A novel formulation and methodology for solving the sampling-based exploration-planning problem - what we in \eqref{eq:realproblem} and \eqref{eq:approx} denote as the momentary Next-Best-Trajectory problem. This methodological difference between ERRT and other exploration-planners lead us to the other contributions that relate to solving that problem. Here, candidate goals of certain criteria are generated and connected to a sampled RRT, while only branches that reach those goals are evaluated for distance, information gain, and predicted actuation. This limits the needed computational effort as only a set number of branches must be evaluated, which is the basis for the following two contributions as they are computationally heavy.
    \item We combine iterative path improvements with the computation of model-based actuation through solving an NMPC problem to refine trajectories and to enable the evaluation of the actuation cost along the trajectory, which is, to the authors best knowledge, a concept only tested in ERRT. 
    \item The ERRT structure allows for the evaluation of information gain \textit{along} the candidate branches, not only at the frontiers or end-goals. This enables a more proper consideration of information-gain prediction during the branch selection step, while the ERRT implementation still allows for low computation times as is needed for a real-time implementation. 
\end{itemize}

The main improvements and extensions of the framework that are included in this manuscript focus on the implementation, evaluations, comparisons, and the code release. They can be stated as:

\begin{itemize}
    \item A full integration of the framework with the state-of-the art occupancy mapper UFOmap~\cite{duberg2020ufomap}, that explicitly models the unknown space in addition to free and occupied. This allows for a proper consideration of a continuously generated map of the explored area, and for the computation of information gain through a 3D LiDAR model and the representation of unknown space. It also enables the integration of ERRT to hardware sensor data input (3D pointcloud scans).
    \item An upgraded sampling method and path improvement framework that generates efficient and robot-safe paths through the environment by representing the UAV as a volumetric object when checking for collisions with the environment. As will be demonstrated, our RRT$^*$-based tree generation enables local robot-safe tree expansion and relevant path extraction in complex 3D environments and can be directly tied to the robot exploration problem, while executed on limited hardware. 
    \item Significant simulation evaluations in large-scale subterranean environments of multiple kinds from large caves to narrow tunnels, to urban indoor areas. Comparisons are performed with two state-of-the-art methods in complex and challenging simulation worlds.
    \item Field Evaluations in narrow and constrained subterranean tunnels towards fully integrating the ERRT framework in the context of a hardware implementation and in conjunction with supportive modules that enable the full autonomy.
\end{itemize}
The result is a fully integrated and realised next-best-trajectory method for agile local exploration for UAVs for completely unknown and unstructured environments. The method can take the path distance, model-based actuation, and predicted information gain into account when selecting the next trajectory to travel along. Additionally, the implementation code for ERRT is released open-source for the community to deploy and use at \url{https://github.com/LTU-RAI/ExplorationRRT}.


\section{Implementation} \label{sec:implementation}
\subsection{Overview}
The implementation of ERRT follows a multi-stage approach where the various components in \eqref{eq:approx} are calculated in sequence and will be explained in more detail in the following Sections \ref{sec:goal} through \ref{sec:evaluation}. The stages can be summarized in order as: (1) Candidate Goal Generation, (2) Robot-safe RRT$^*$ tree expansion and path extraction, (3) Computation of model-based trajectories and actuation, 4) Computation of information gain and trajectory evaluation

The tree-based exploration process is applied only on a local subset of the whole map space as RRT-based solutions lose their efficiency relatively quickly for larger areas. As such $V_\mathrm{map}$ is replaced with local map $V_\mathrm{map}^l$ in \eqref{eq:realproblem}, which in the implementation implies extracting a subset of the occupancy map confined by bounding box of the desired sampling volume centered on the estimated robot position $\hat{p}$. As such, in this implementation, ERRT on its own should be considered a local exploration-planning framework, but as will be seen in Section \ref{sec:simulation} ERRT still enables the exploration of large-scale subterranean areas. 

An important note is that due to the generation of a set number of discrete candidate goals, the steps (2)-(4) can deploy relatively computationally heavy processes without risking significant computation times; iterative path optimization for each candidate path, calculating the robot actuation along the trajectories as a NMPC problem, and calculating the model-based information gain at each pose $\chi$ in the trajectories without any averaging or simplification. We should be clear that some components closely follow the initial work \cite{lindqvist2021exploration}, but are still included in this manuscript for completeness.

The implementation of ERRT is mainly written in C++ and is based on the Robot Operating System (ROS) \cite{quigley2009ros}. The algorithm heavily uses the UFOmap\cite{duberg2020ufomap} library for efficient volumetric collision checks for robot-safe path generation and information gain calculations (number of unknown voxels in sensor view). Additionally, the NMPC module for model-based actuation computation uses the Optimization Engine~\cite{sopasakis2020open} as it provides lighting-quick solutions for actuation-trajectories while enabling the use of a nonlinear kinodynamic model of the UAV (as opposed to, for example, a simple double-integrator model) while also applying model-based constraints.

\subsection{The Local Map Sampling Space}
As previously mentioned, ERRT operates as a local planner to leverage the efficiency of the RRT tree-expansion. In the code implementation, the local map $V^l_\mathrm{map}$ is described by a bounding box centered on the estimated robot position $\hat{p}$. The bounding box size is set by an input parameter to the algorithm that defines the volume that is centered on $\hat{p}$ and stretches to the edge of $V^l_\mathrm{map}$ e.g. the parameter describes the desired side length of the bounding box $V^l_\mathrm{map}$.  $V^l_\mathrm{map}$ is then further constrained by the maximal and minimal $x,y,z$-coordinates of free voxels $F^l \in V^l_\mathrm{map}$ to limit unnecessary sampling in completely occupied or unknown areas. In general, the local map space is used to constrain the sampling volume for randomly generated points (or voxels) $P$ used throughout the algorithm, and when later sections mention $V^l_\mathrm{safe}$, it implies a subset of $V^l_\mathrm{map}$ where a volumetric collision check has ensured that $P$ lies a distance at least $r_{robot}$ from any occupied or unknown voxels. When the ERRT algorithm is applied for simulations and experiments later in the manuscript, $V^l_\mathrm{map}$ is updated at every call to the algorithm e.g. for each new exploration trajectory.

\subsection{Candidate Goal Generation} \label{sec:goal}
For any sampling based method to robot exploration a critical parameter is how to sample candidate solutions. In ERRT this process starts by generating candidate goal poses $g^\mathrm{c}$, which based on \eqref{eq:approx} are $n_\mathrm{traj}$ in number, all have $\nu(g^\mathrm{c}) > 0$ (e.g. guaranteeing that regardless of the path $\bm{\chi}_j$ to $g^\mathrm{c}_j$ the information gain is nonzero for the full path), and all lie in the local safe space $g^\mathrm{c} \in V^l_\mathrm{safe}$. Additionally, to ensure an efficient spread of candidate goals we apply \textit{Poisson disk sampling} such that all candidate goals are a set distance $d_{g^\mathrm{c}}$ from each other, and a set distance from the robot position $\hat{p}$. 
The evaluation of information gain is more appropriately described in \ref{sec:evaluation}, but in short to ensure $\nu(g^\mathrm{c}) > 0$ a volumetric check is made on a sampled point $P$ based on a 3D LiDAR field-of-view with sensor range $S_r$ and vertical field-of-view $S_\theta$ such that $\nu(P)$ returns the number of unknown voxels $\{U\}$ that are in field-of-view and not blocked by any occupied voxels $\{O\}$. In the case of goal generation, the check stops if one $\{U\}$ is found for efficiency. 
If all such conditions are met, the sampled point is added as a candidate goal $g^\mathrm{c}$. Additionally, although $\nu(g^\mathrm{c}) > 0$ is the condition for continued exploration, the code implementation allows the user to select also a $\nu(g^\mathrm{c}) > \nu_\mathrm{min}$ but this comes at a cost of computational efficiency as total information gain must be calculated for each sampled goal that goes through the initial check  $g^\mathrm{c} \in V^l_\mathrm{safe}$.
\subsection{RRT-Structure \& Path Optimization\label{sec:RRT}}
While there are many modern improved versions of RRT tree-expansion~\cite{wang2020neural,otte2016rrtx,qi2020mod}, ERRT is based on relatively standard RRT$^*$ structure. RRT$^*$ has been used extensively~\cite{noreen2016optimal}, and as such the core method will not be extensively described in this article. The following text can be more compactly described by Algorithm \ref{alg:tree_generation} where all related variables, motivations, and descriptions can be found in the following text.

\begin{algorithm}[ht] 
\SetAlgoLined
\textbf{Inputs:} $g^c,V^l_\mathrm{map}, \hat{p}$  \\
\KwResult{Generate Robot-safe Candidate Branches}
\While{$size(\bm{N}) \leq N_\mathrm{max}$}{ 
    P = $\mathrm{generate\_random}(V^l_\mathrm{map}$) \\
    \If{$P \in V^l_\mathrm{safe}$}{
         $N_\mathrm{closest} \leftarrow \mathrm{find\_closest}(P,\bm{N})$\\
        \If{$P \rightarrow N_\mathrm{closest} \in V^l_{safe}$}{$\bm{N} \leftarrow \mathrm{add\_leaf}(N_\mathrm{closest},P)$}}}
\While{$j \leq n_{traj}$}{$N_\mathrm{extend} \leftarrow \mathrm{extract\_branches}(g^c_j, \bm{N}, d_\mathrm{extend})$ \\
\For{$\bm{\mathrm{all}} \hspace{0.1cm} N_\mathrm{extend}$}{\If{$g^c_j \rightarrow N_\mathrm{extend,i} \in V^l_\mathrm{safe}$}{$\bm{N}\mathrm{\leftarrow add\_leaf(N_\mathrm{extend,i}, g^c_j})$}} 
$\bm{N}^{g^c} \leftarrow \mathrm{find\_shortest}(g^c_j, \bm{N})$ \\
\If{$N^{g^\mathrm{c}}_{j,1} \rightarrow N^{g^\mathrm{c}}_{j,n_j} \in V_\mathrm{safe}^l$}{$N^{g^\mathrm{c}}_{j} \leftarrow \mathrm{linspace(simplify(N^{g^\mathrm{c}}_{j}))}$ \\
$\bm{N}^{g^c}_\mathrm{reduced} \leftarrow \mathrm{add\_branch}(N^{g^\mathrm{c}}_{j})$}
\Else{$N^{g^\mathrm{c}}_{j} \leftarrow \bm{\mathrm{try}} \hspace{0.1cm} \mathrm{improve\_path}(N^{g^\mathrm{c}}_{j}, recursive)$ \\
$N^{g^\mathrm{c}}_{j} \leftarrow\bm{\mathrm{try}} \hspace{0.1cm} \mathrm{improve\_path(\mathrm{linspace}}(N^{g^\mathrm{c}}_{j}), rec)$ \\
$\bm{\mathrm{if\hspace{0.1cm} try \hspace{0.1cm} failed}} \hspace{0.1cm} N^{g^c}_j \leftarrow \mathrm{linspace}(N^{g^\mathrm{c}}_{j})$ \\
$\bm{N}^{g^c}_\mathrm{reduced} \leftarrow 
\mathrm{add\_branch}(N^{g^\mathrm{c}}_{j})$}
j++} \textbf{Output:} $\bm{N}^{g^c}_\mathrm{reduced}$
\caption{Tree Expansion \& Path Improvement}\label{alg:tree_generation}
\end{algorithm}

Assume that we initialize the tree $\bm{N}$ with a root node at the robot current position $\Hat{p} = [p_\mathrm{x},p_\mathrm{y},p_\mathrm{z}]$.
In our implementation, random points $P = [P_\mathrm{x}, P_\mathrm{y}, P_\mathrm{z}]$ are generated in the local sampling space $V_\mathrm{map}^l$. $P$ is checked against its occupancy state in $V_\mathrm{map}^l$ first to ensure that it is in $V_\mathrm{safe}^l$ by performing a volumetric spherical collision check with robot radius $r_\mathrm{robot}$ centered on $P$. Then, we search through $\bm{N}$ to find the closest node $N_\mathrm{closest} \in \bm{N}$ in the tree, and check if the straight line $P\rightarrow N_\mathrm{closest} \in V_\mathrm{safe}^l$ by collision-checking the cylindrical volume between $N_\mathrm{closest}$ and $P$ with radius $r_\mathrm{robot}$. If all checks pass, $P$ is added as a child node to $N_\mathrm{closest}$ in the complete tree $\bm{N}$, and the process is repeated. We found that adding $P$ to $\bm{N}$ directly as opposed to the more usual "step" from $N_\mathrm{closest}$ towards $P$ was more efficient for searching a smaller local space with fewer iterations in combination with the subsequent path improvement step. After a set number nodes $N_\mathrm{max}$ have been added to $\bm{N}$ we compute which branches in $N_\mathrm{extend} \in \bm{N}$ can be extended by the candidate goals $g^\mathrm{c}$ by checking all nodes and goals for the conditions $N \rightarrow g^\mathrm{c} \in V_\mathrm{safe}^l$, and $\mid N - g^\mathrm{c} \mid < d_\mathrm{extend}$  where $d_\mathrm{extend}$ denotes the maximum distance of nodes in $N$ to $g^c$ where the connection attempt is performed as to limit the number of nodes to test for. We then extract the shortest such branches to each goal node and we can denote them as $\bm{N}^{g^\mathrm{c}}_j$ where $j = 1,2 .... n_\mathrm{traj}$. As such the total search tree $\bm{N}$ has been reduced to a set of $n_\mathrm{traj}$ candidate branches that have guaranteed information gain in them since $\nu(g^\mathrm{c}_j) > 0$.

A common problem in sampling-based (especially RRT) solutions to real-time path planning is that the resulting paths are not always optimally short as we are forced to limit the number of iterations due to computation time limits. We found it more efficient to construct a more sparser tree $\bm{N}$ and then try to optimize the resulting $\bm{N}^{g^\mathrm{c}}$ through iterative volumetric collision checks.
The method is fundamentally based on checking if $\bm{N}^{g^\mathrm{c}}$ can be reduced by removing redundant nodes. Let's denote each $\bm{N}^{g^\mathrm{c}}_j = [N^{g^\mathrm{c}}_{j,1}, N^{g^\mathrm{c}}_{j,2}.....N^{g^\mathrm{c}}_{j,{n_j}}]$, where $n_j$ denotes the number of nodes in $\bm{N}^{g^\mathrm{c}}_j$ and as such $N^{g^\mathrm{c}}_{j,{n_j}} = g^\mathrm{c}_j$. 
The program starts by checking if $N^{g^\mathrm{c}}_{j,1} \rightarrow N^{g^\mathrm{c}}_{j,n_j} \in V_\mathrm{safe}^l$ e.g. if the straight line from the robot position $\Hat{p}$ to the candidate goal $g^\mathrm{c}_j$ is collision-free and safe. If it is, as the straight line is the shortest possible path, all other nodes are redundant and are removed. If not, the process continues with checking if $N^{g^\mathrm{c}}_{j,2} \rightarrow N^{g^\mathrm{c}}_{j,n_j} \in V_\mathrm{safe}^l$ etc. If a path improvement is found between index $i$ and the end node, the process is restarted with checking $N^{g^\mathrm{c}}_{j,1} \rightarrow N^{g^\mathrm{c}}_{j,i} \in V_\mathrm{safe}^l$ and so on. This is implemented in a recursive way such that if a path improvement is found, nodes are removed and the function attempts to find another path improvement on the new branch. This step is limited by a maximum time duration for finding improvements.
We can denote the resulting tree as $\bm{N}^{g^\mathrm{c}}_\mathrm{reduced}$. The path shortening step is then repeated but with added nodes at a set distance from each other in the resulting $\bm{N}^{g^\mathrm{c}}_\mathrm{reduced}$, as an additional possibility at connecting the nodes and making branches with a smaller total path length. This last step is mainly focused on making smoother paths, and to shorten paths around corners or obstacles. 

Of course this is a computationally heavy process, but we are only checking $n_\mathrm{traj}$ solutions in total, not every branch in $\bm{N}$. The end-result are the $n_\mathrm{traj}$ optimized (shortened) candidate graphs/branches, with a specified distance between nodes, where each branch has its end-node at the candidate goal position. For the sake of notation in the following Section \ref{sec:actuation} let us denote them as the reference position trajectories $\bm{X}^\mathrm{ref}_j = [X^\mathrm{ref}_{j,1},X^\mathrm{ref}_{j,2}...X^\mathrm{ref}_{j,n_j}]$.
\subsection{Actuation Trajectory\label{sec:actuation}}
This section will describe how ERRT solves for model-based trajectories and calculates the actuation vector $\bm{u}$ required to follow the reference trajectories $\bm{X}^\mathrm{ref}_j$. The reason for adding a consideration of the robot actuation into the exploration-planning problem are threefold: 1) Paths with high actuation will be harder for the robot to follow (and require more energy to do so) and can induce unwanted controller behavior, 2) UAVs and other robotic platforms are dynamically constrained in their motion, and the generated trajectories should reflect that, 3) Continued forward exploration behavior is preferred over constant backtracking or sharp direction changes. In other works on robot exploration this behavior is hard-coded as part of frontier selection~\cite{patel2023ref},  is part of the acceleration sampling~\cite{dharmadhikari2020motion}, or due to "carrot-chasing" approaches to robot exploration~\cite{lindqvist2022compra}. In ERRT that behavior is instead a result of the fact that direction changes requires higher actuation (e.g. high actuation trajectories can still be selected but only if the revenue gain is also higher). We also solve for actuation trajectories as a NMPC problem (following the approach in previous works~\cite{sathya2018embedded, small2019aerial, lindqvist2020dynamic}), that fully considers the nonlinear model of the UAV and its dynamical constraints. The nonlinear model of the UAV can be written as: 
\begin{subequations}
\small
\label{eq:mavkinematic}
\begin{align}
        \dot{p}(t) &= v(t) \\ 
        \dot{v}(t) &= R(\phi,\theta) 
        \begin{bmatrix} 0 \\ 0 \\ u_\mathrm{T} \end{bmatrix} + 
        \begin{bmatrix} 0 \\ 0 \\ -g \end{bmatrix} - 
        \begin{bmatrix} A_x & 0 & 0 \\ 0 &  A_y & 0 \\ 0 & 0 & A_z \end{bmatrix} v(t), \\ 
        \dot{\phi}(t) & = \nicefrac{1}{\tau_{\phi}} (K_\phi u_\phi (t)-\phi(t)), \\ 
        \dot{\theta}(t) & = \nicefrac{1}{\tau_{\theta}} (K_\theta u_\theta(t)-\theta(t)).
\end{align}
\end{subequations}
We can define the state-vector as $\chi = [p,v,\phi, \theta]$ consisting of position, velocity, pitch angle, and roll angle states, with the input actuation to the system being $u = [u_\mathrm{T}, u_\phi, u_\theta]$ as the thrust and reference angles in pitch and roll. The acceleration of the system is defined to be the (nonlinear) rotation $R(\phi(t),\theta(t)) \in \mathrm{SO}(3)$ of the thrust vector produced by the UAV motors, as well as the acceleration due to gravity $g$ and linear damping terms $A_x, A_y, A_z$. The roll $\theta$ and pitch $\phi$ states are modelled as first-order systems actuated by $u_\phi, u_\theta$ as to estimate the behavior of an onboard attitude controller acting on those control inputs (and as such also captures that the system will not react instantly to desired accelerations), where $\tau_{\phi}, \tau_{\theta}$ are the time constants and $K_\phi, K_\theta$ are the gains. The system dynamics are discretized with sampling time $\delta_\mathrm{t}$ by the forward Euler to obtain the predictive form $\chi_{k+1} = \zeta(\chi,u)$. Let's also denote the prediction horizon $N$ and that $k+l\mid k$ implies a prediction $l$ time steps forward produced at time step $k$. 
We can now formulate the objective function $J(\bm{\chi}_{k}, \bm{u}_{k})$ as:
\begin{multline}
\label{eq:costfunction}
J(\bm{\chi}_{k}, \bm{u}_{k}) = \sum_{l=0}^{N}   \| X^\mathrm{ref}_l-\chi_{k+l{}\mid{}k}\|_{Q_X}^2
\\
+  \| u_{\mathrm{ref}}-u_{k+l{}\mid{}k}\|^2_{Q_u}
+  \| u_{k+l{}\mid{}k}-u_{k+l-1{}\mid{}k} \|^2 _{Q_{\Delta u}}.
\end{multline}
where $Q_x, Q_u, Q_{\Delta u}$ are positive definite weight matrices for the states, inputs and input rates respectively, and $u_\mathrm{ref} = [g,0,0]$ representing the steady-state input. Let us also right away define constraints on the control inputs $u_{\min} \leq u_{k+l\mid k} \leq u_{\max}$ to realistically restrict the thrust and attitude commands based on the utilized UAV platform's dynamical constraints. 
We can here see that we feed in the optimized RRT-branches $\bm{X}^\mathrm{ref}$ as state references along the horizon (with velocity and angle references set to zero) and output the predicted and dynamically constrained state trajectory $\bm{\chi}$ and the actuation vector $\bm{u}$. The resulting optimization problem can be written as:

 \begin{subequations}\label{eq:nmpc}
\begin{align}
    \operatorname*{Minimize}_{
        \bm{u}_k, \bm{\chi}_k
    } \,
    & J(\bm{\chi}_{k}, \bm{u}_{k})
    \\
    \text{s. t.:}\,& 
    \chi_{k+l+1\mid k} = \zeta(\chi_{k+l\mid k}, u_{k+l\mid k}),\notag
     \\ & l=0,\ldots, N,
    \\
    &u_{\min} \leq u_{k+l\mid k} \leq u_{\max},\, l=0,\ldots, N, \\
    &\chi_{k\mid k} {}={} \hat{x}_k.
\end{align}
\end{subequations}
where the initial state $\chi_{k\mid k}$ is the full UAV state $\Hat{x}$ estimated by onboard state-estimation. 
The problem is solved by the nonlinear nonconvex parametric optimization software Optimization Engine~\cite{sopasakis2020open} and the PANOC~\cite{stella2017simple} algorithm. The NMPC problem is solved for all reference trajectories $\bm{X}^\mathrm{ref}_j$ to generate the optimally actuated (based on objective function \eqref{eq:costfunction}) trajectories $\bm{\chi}_j$ and actuation vectors $\bm{u}_j$ that track $\bm{\chi}_j$.

A clear limitation is that the prediction horizon is pre-defined for the problem, and as such we can only compute the actuation for a limited number of nodes in $\bm{X}^\mathrm{ref}$. But, reasonable values for the branch step size in the branch (and the associated sampling time in the NMPC $\delta_t$) coupled with the local sampling space $V_\mathrm{map}^l$ mean that this limitation very rarely comes up. If $n_j > N$ we will simply have to use the original $\bm{X}^\mathrm{ref}$ for the last indices. 

An added benefit is that by relating the tree-branch sampling distance to the sampling time $\delta_t$ we can generate actuation trajectories for a certain exploration speed, and since $\bm{\chi}$ is a full-state trajectory we can provide an onboard reference tracking controller not only with position references $p_\mathrm{ref}$ but also velocity references $v_\mathrm{ref}$ (and technically also angle references) that take the dynamics of the system into consideration (ex. slow down before a sharp turn) and can smoothen the transition between position references in the trajectory. Inherently we will also have a measure for faster exploration resulting in higher actuation costs. 

As an example for the reader: for a prediction horizon of $N = 50$, a $\delta_t = \unit[0.4]{s}$, and a branch step size of $\unit[0.4]{m}$ (e.g. we generate trajectories with an average exploration speed of $\unit[1]{m/s}$), and $n_\mathrm{traj} = 50$ sampled solutions, the computation time of the NMPC problem was averaging $\unit[3]{ms}$ per trajectory resulting in a $\unit[0.15]{s}$ total increase in computation time resulting from the NMPC problems. The actuation trajectory can then also cover $\unit[20]{m}$ which is a reasonable and relatively large size for the side of the bounding box that defines the local sampling space $V_\mathrm{map}^l$. 

The resulting actuation-based trajectory $\bm{\chi}_j$ is the trajectory used in the following trajectory evaluation (Section \ref{sec:evaluation}) and the optimal selected one is the final output of ERRT.

\subsection{Trajectory Evaluation\label{sec:evaluation}}
After the $n_\mathrm{traj}$ candidate actuation-based trajectories $\bm{\chi}_j$ are computed, the final step is evaluating them according to the minimization in \eqref{eq:approx} in order to select the next-best-trajectory $\bm{\chi}^*$.
\subsubsection{Information Gain Calculation}
Calculating the information gain along $\bm{\chi}_j$ requires a sensor model. In this work, we use a simple 3D LiDAR model with only two parameters: the sensor range $S_\mathrm{r}$ and the vertical field-of-view angle $S_\theta$. The UFOmap library enables us to extract all unknown voxels $\{U\}$ in a LiDAR field-of-view geometry from the robot position or from any point in the trajectory. That subset of unknown voxels $\{U\}_\mathrm{fov}$ in $V^l_\mathrm{map}$ can then be analyzed if they are in line-of-sight of the sensor e.g. that $\chi_{j,i} \rightarrow \{U\}_\mathrm{fov} \in V_\mathrm{free}^l$ for each $\{U\}_\mathrm{fov}$. We extract the geometry and perform the collision checks with $S_\theta$ vertical field-of-view, range $S_\mathrm{r}$, and with the center point at a certain position in the trajectory, to mimic the predicted field-of-view of a 3D LiDAR along the trajectory and to extract all unknown voxels within that field-of-view. We want to perform such checks along the trajectory, but doing so at each point in $\bm{\chi}_j$ would be too computationally expensive and would have extreme overlap as the distance between nodes is much smaller than the sensor range. As such, the ERRT implementation performs these checks 
not at every node in $\chi$, but only at nodes that are a specific distance apart. This set distance is a tuning parameter and input to the algorithm we can define as $d_\mathrm{info}$. During this process, only $\{U\}_\mathrm{fov}$ that were not seen at a previous node in the branch are tested for the condition $\chi_{j,i} \rightarrow \{U\}_\mathrm{fov} \in V_\mathrm{free}^l$ to remove duplicates.
We then denote the resulting total non-duplicate number of $\{U\}$ along $\bm{N}^{g^\mathrm{c}}_\mathrm{reduced,j}$ as the information gain for that trajectory. This is a simplification step, but will produce a better approximate of predicted information gain as opposed to only evaluating, for example, the edge nodes in the full tree/graph $\bm{N}$ which would also be much more computationally expensive, or evaluating information gain only at certain selected frontier locations. In the ERRT program the equivalent would be only evaluating information gain at each $g^c$, which although offers a less accurate prediction of explored volumes, can be enabled as an input parameter in the ERRT implementation for a reduced computational complexity of the program.

The information gain revenue is calculated as:
\begin{equation}
    C_\mathrm{i}(\nu(\bm{\chi}_j)) = K_\mathrm{i} \nu(\bm{\chi}_j),
\end{equation}
where $K_\mathrm{i}$ denotes a gain related to the relative emphasis on the information revenue parameter. As such $C_\mathrm{i}(\nu(\bm{\chi}_j))$ represents the predicted sensor-based information gain along the trajectory.
\subsubsection{Distance \& Actuation}
Evaluating the distance cost $C_\mathrm{d}(\bm{\chi}_j)$ is done by simply summing the distance between consecutive positions in the actuation-based trajectories $\bm{\chi}_j$ as: 
\begin{equation}
    C_\mathrm{d}(\bm{\chi}_j) = K_\mathrm{d} \sum^{n_j}_{i=2} \mid \mid \chi_{j,i} - \chi_{j,i-1} \mid \mid 
\end{equation}
with $K_\mathrm{d}$ denoting the relative emphasis on the length of the trajectory. Finally, computing the actuation cost is done by feeding the actuation vector $\bm{u}_j$ back into \eqref{eq:costfunction} as: 
\begin{multline}
\label{eq:actuationcost}
C_\mathrm{u}(\bm{u}(\bm{\chi}_j)) = K_\mathrm{u}\sum_{i=0}^{N} \| u_{\mathrm{ref}}-u_{j,i}\|^2_{Q_u} +\| u_{j,i}-u_{j,i-1} \|^2 _{Q_{\Delta u}}.
\end{multline}
The resulting $C_\mathrm{u}(\bm{u}(\bm{\chi}_j))$ represents the (nonlinear) model-based actuation cost along each full trajectory $\bm{\chi}_j$. With all trajectories generated in $V^l_\mathrm{safe}$ and all costs computed, the ERRT program can now evaluate the expression in \eqref{eq:approx} to find the approximate next-best-trajectory $\bm{\chi}^*$.

\section{Simulation Results} \label{sec:simulation}
The best way to view the efficiency and real-time behavior of the algorithm is through the simulation evaluation video which can be found at \url{https://youtu.be/R2J-dpVL57M}. In this video we showcase ERRT in two different simulation worlds of wide interconnected subterranean caves and tunnels as well as an urban hospital world with corridors, narrow doors, and rooms. In this video we visualize the generated trajectories and the predicted exploration along them. We also highlight the efficient trajectory transitions and fast re-calculation. The ERRT tuning parameters of greater importance used for the simulations can be found in Table \ref{table:sim_param}. We target relatively rapid re-calculation and greedy behavior. We note that the sensor range $S_r$ is set quite low. We have found that artificially lowering the sensor range both greatly reduces computation for information gain checks, and promotes the UAV to move deeper into unknown areas. The sensor range for occupancy mapping is still set higher to match the performance of the onboard LiDAR.
\begin{table}[!ht]
\begin{center}
\caption{Critical ERRT parameters used for simulations.}
 \label{table:sim_param}
\begin{tabular}{ |c|c|c|c|c|c|c|c| } 

 $V^l_\mathrm{map}$ & $n_\mathrm{traj}$ &  $S_r$ &$size(\bm{N})$ & $d_\mathrm{info}$ & $K_d$ & $K_i$ & $K_u$ \\ 
 40m & 60 & 10m & 2000 & 6m &0.3 & 0.4 & 0.1 \\ 
\end{tabular}
\end{center}
\end{table}

In these simulations, ERRT is coupled with a full-state trajectory tracking controller and to simulate the UAV itself, its onboard sensor, and its dynamics and attitude control we use the RotorS framework~\cite{furrer2016rotors}. The simulated UAV is equipped with a 45$^\circ$ vertical field-of-view Ouster 32-beam 3D LiDAR. The robot localization is here provided by the simulator and not by SLAM, and as such the all simulations and comparisons operate under zero localization noise or uncertainty.
As ERRT is focusing on efficient information-gain maximizing behavior, is designed to be a local exploration module, and since real UAVs have limited flight time, we set up simulation runs in each world where we let the UAV explore for a set number of minutes at realistic navigation speeds of $\unit[0.8-1]{m/s}$, and as such the main goal should be seen as maximizing efficient local exploration in a set time frame not complete coverage of the area. In the following simulated exploration runs the desired safety distance $r_{robot}$ is set to $\unit[0.3]{m}$, only slightly above the size-radius of the simulated vehicle. As the simulation set-up means operating under perfect localization and control, we do not have to artificially inflate the safety distance significantly above the size-radius of the robot, which does help it to pass through the more narrow areas in these environments.

\subsection{Subterranean Exploration}\label{sec:sim_exploration}
The framework was extensively tested in the Gazebo simulation software in large-scale subterranean maps from the aforementioned DARPA Robotics Challenge, that present two very different types of environments; 1) The DARPA SubT Final Stage World - a Gazebo world mimicking the exact layout of the final stage of the competition with narrow tunnels, entrances, junctions, and obstacle-rich areas, and 2) The DARPA SubT Cave World with 10-15m wide tunnels and caves, 3D layouts and shafts, and in general very large-scale structures. In this section, while we include simulations from both worlds, the main focus will be on the narrow environments while the large-scale caves and voids will be further analysed in the comparison section \ref{sec:comparison}. As the official documentation of these Gazebo worlds, and others, are now discontinued we provide a repository at: \url{https://github.com/LTU-RAI/darpa_subt_worlds.git}.

\begin{figure}[!htbp]
    \centering
\includegraphics[width=\linewidth]{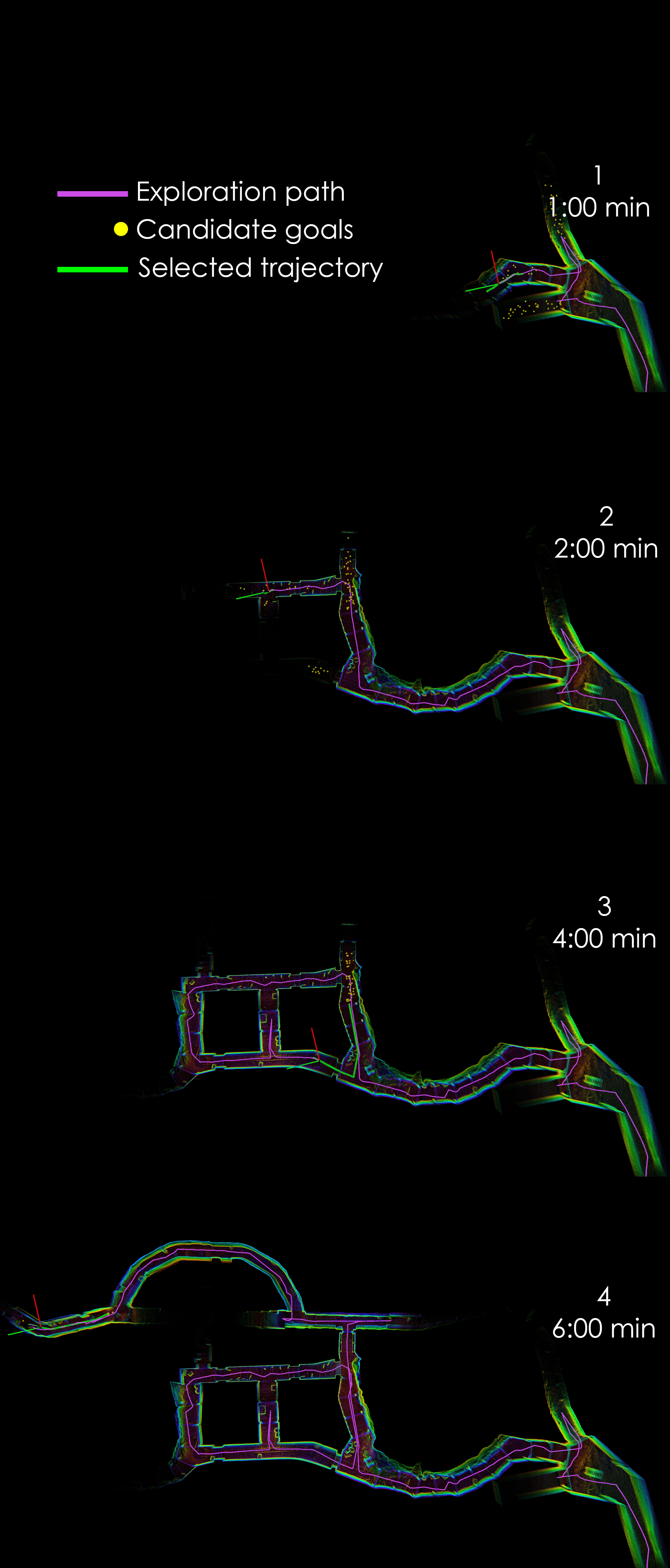}
    \caption{Local Exploration over 6 minutes in the DARPA Final Stage gazebo world - mimicking the conditions of the real competition. Figures highlighting the progress of exploration at set times. Total exploration path length was around $\unit[250]{m}$.}
    \label{fig:final_progress}
\end{figure}

Figure \ref{fig:final_progress} shows the progression of exploration for a 6 minute mission in the DARPA Final Stage World of extremely narrow tunnels filled with obstacles. Here, the ERRT-guided UAV can be seen exploring the Final Stage environment at certain set time stamps throughout the mission. The exploration run starts in the bottom-right of the sub-maps at an entrance to the underground area and continues through a complex junction-filled area of tunnels. The UAV explores the tunnels with minimal backtracking and unwanted maneuvering, while continuously exploring new unknown areas.
\begin{figure}[!htbp]
    \centering
\includegraphics[width=\linewidth]{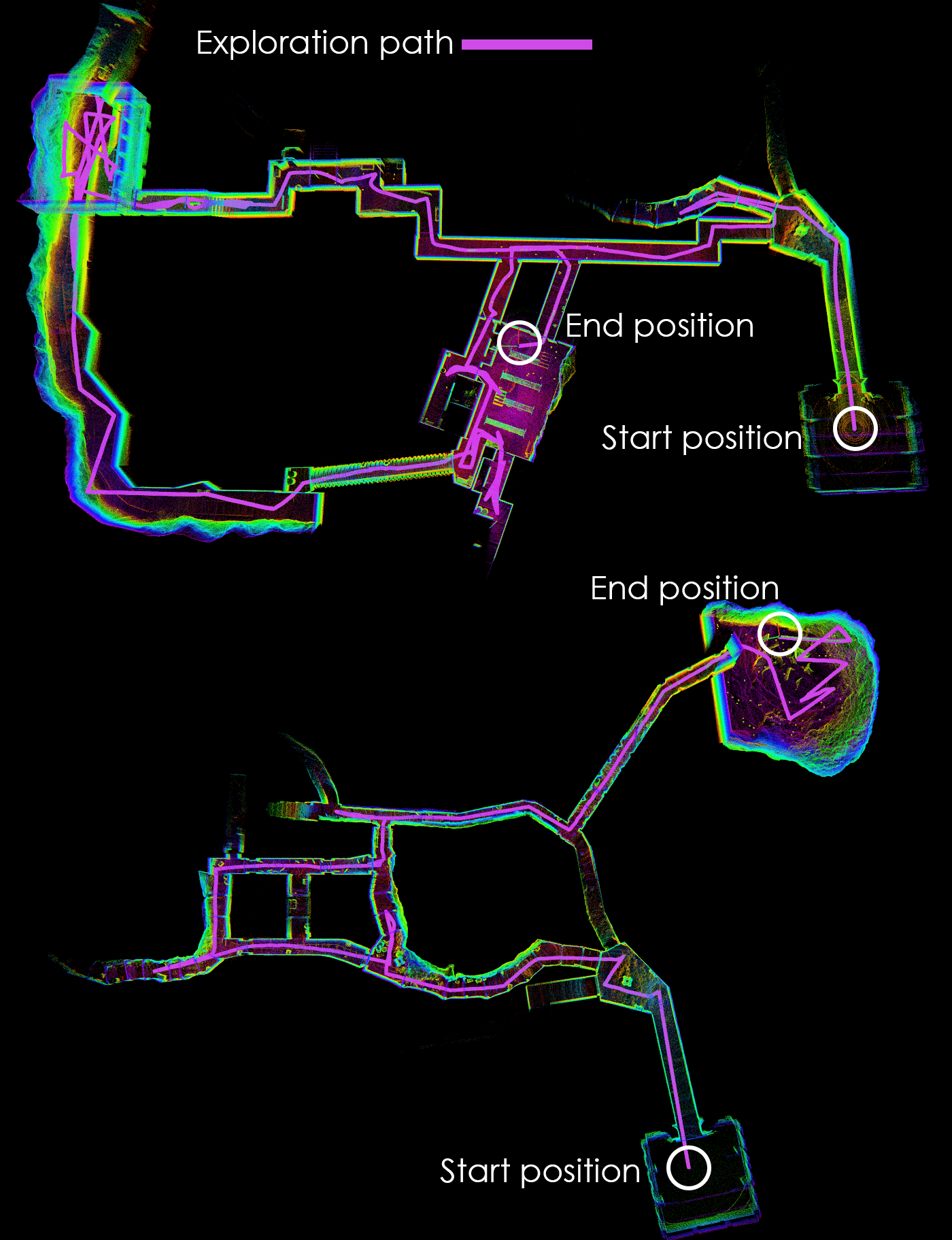}
    \caption{Two more ERRT exploration runs into different parts of the area, with varying kinds of environments from urban warehouse like areas to tunnels and caves.}
    \label{fig:final_moremaps}
\end{figure}
Figure \ref{fig:final_moremaps} shows two more ERRT runs going into other sections of the world with open voids and warehouse-like rooms.
\begin{figure}[!htbp]
    \centering
\includegraphics[width=0.9\linewidth]{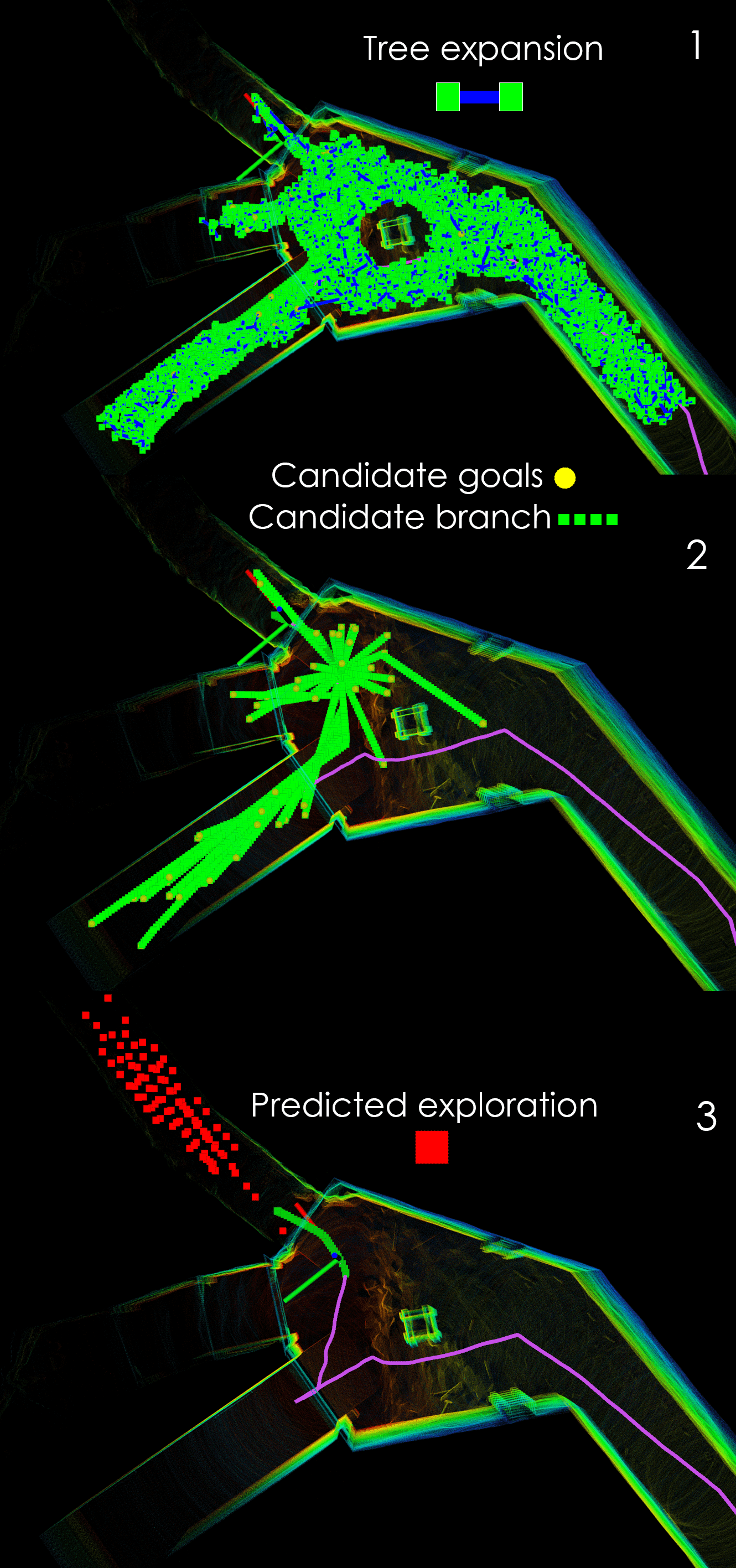}
    \caption{The ERRT process - 1) local robot-safe Tree Expansion filling $V^l_\mathrm{safe}$, 2) pseudo-random goal sampling $g^\mathrm{c}$ (yellow dots) and improved actuation-paths $\bm{\chi}_j$ (green), and 3) selected path $\bm{\chi}^*$(green) with marked unknown voxels that will be discovered along the "next-best-trajectory" (red) (bottom).}
    \label{fig:errt_stages}
\end{figure}
Figure \ref{fig:errt_stages} shows the general ERRT concept and program; 3D robot-safe tree expansion in the local robot-safe space $V^l_\mathrm{safe}$, the computation of improved (shortened) actuation trajectories $\bm{\chi}_j$ extended to the sampled $g^\mathrm{c}$  with $\nu(g^\mathrm{c}) > 0$, and finally the selected  $\bm{\chi}^*$ with a visualisation of the predicted explored volume (red markers highlighting which unknown voxels will be in sensor view) that will be achieved by following that trajectory. Not shown is the very first step of generating the candidate goals $g^\mathrm{c}$ prior to tree expansion. 

\begin{figure}[!htbp]
    \centering
\includegraphics[width=0.9\linewidth]{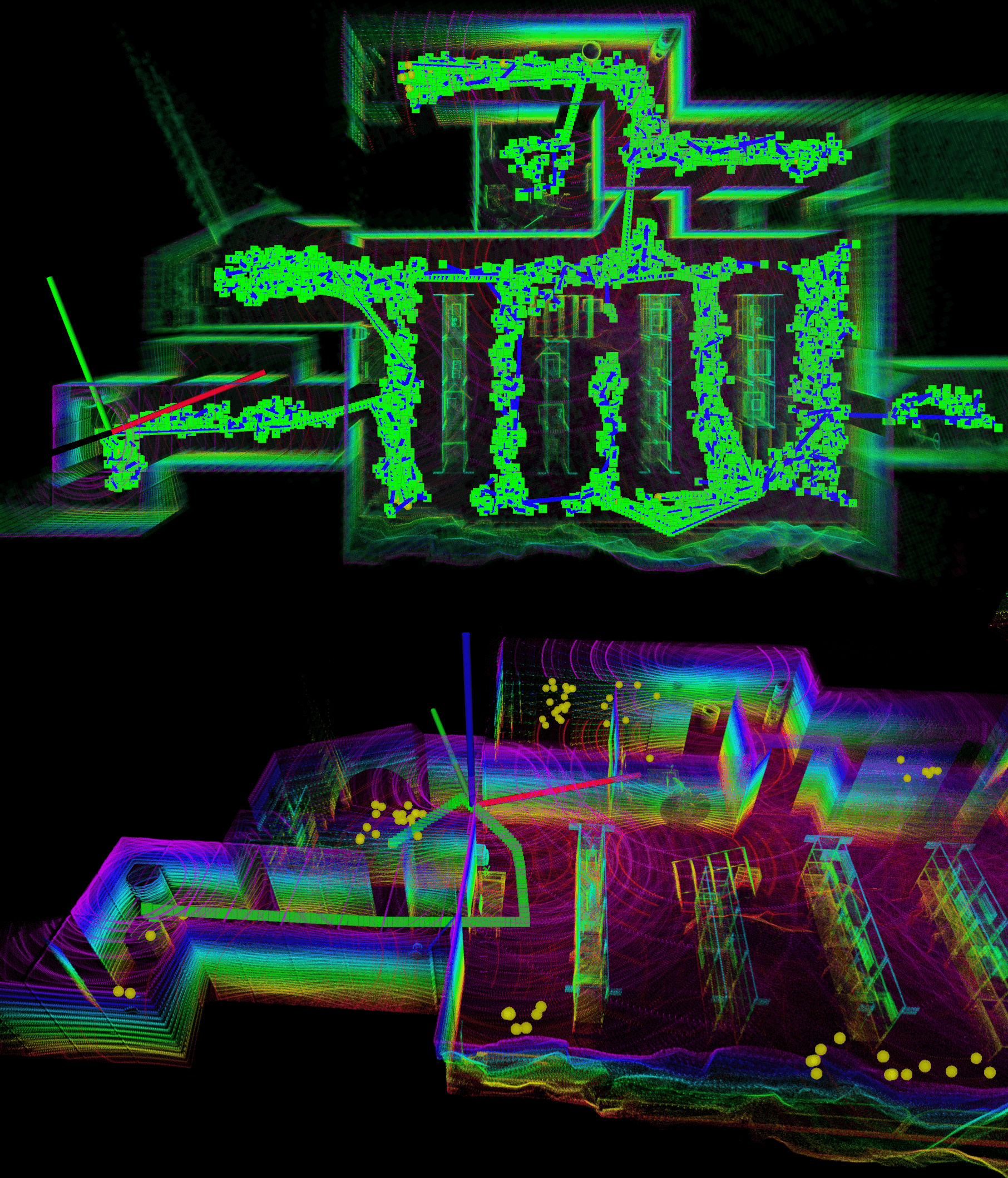}
    \caption{Robot-safe 3D Tree expansion in an obstacle-rich and complex warehouse-like area (top), and selected exploration "next-best-trajectory" (green line) (bottom). The "clumps" of yellow dots are the candidate goals and define the areas of interest to visit as they have information gain.}
    \label{fig:errt_graph}
\end{figure}

Additionally, as a showcase for the sampling-based tree expansion, Figure \ref{fig:errt_graph} shows the robot-safe tree in a complex, multi-room, and obstacle-filled local sampling area at a different location in the same DARPA Final Stage World, as well as the selected exploration path to a new area. We can see that the tree expansion efficiently covers the local sampling space, even in the extremely narrow entrances to various rooms in the area. 

\begin{figure}[!htbp]
    \centering
\includegraphics[width=\linewidth]{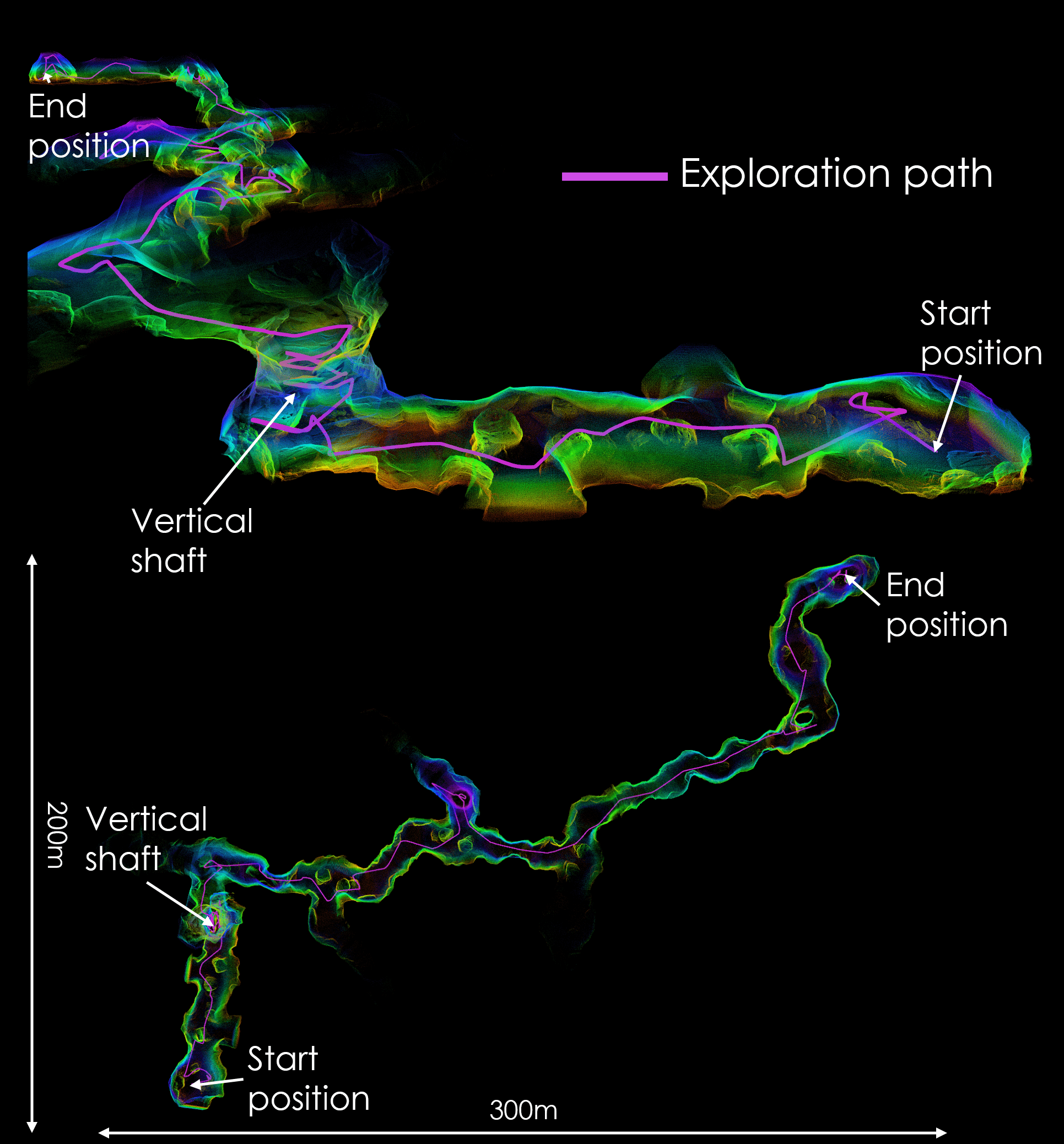}
    \caption{Explored area after 15 minutes in the DARPA Cave World - approx. 10-15m wide tunnels. 3D view highlighting the vertical structures (top) and the map overview (bottom).}
    \label{fig:cave_world}
\end{figure}

\begin{figure}[!htbp]
    \centering
\includegraphics[width=\linewidth]{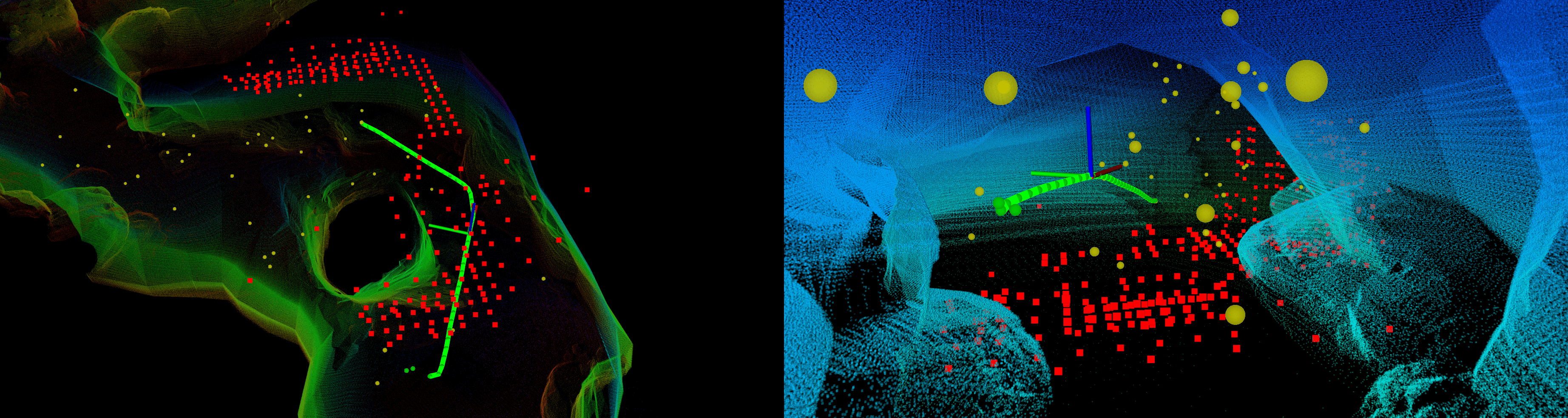}
    \caption{Two separate critical instances of trajectory generation during the wide cave exploration - selection of high-information branches. The predicted information gain (red) along the selected trajectory (green) leading to one of the candidate goals (yellow).}
    \label{fig:cave_snaps}
\end{figure}

\begin{figure}[!htbp]
    \centering
\includegraphics[width=\linewidth]{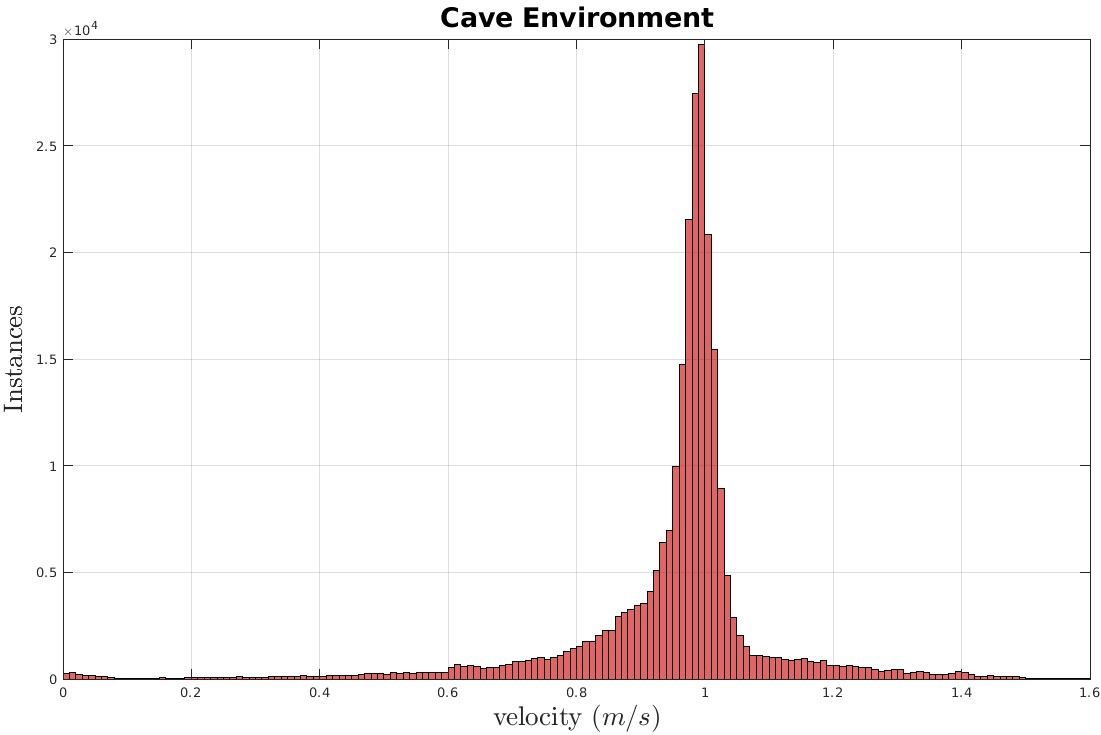}
    \caption{Velocities as a histogram showing how frequent different velocities were measured during the mission. The Figure highlights that ERRT performs efficient trajectory transitions without slowing down.}
    \label{fig:velocity}
\end{figure}

Figure \ref{fig:cave_world} shows a 15 minute exploration run in the DARPA Cave World that features complex and wide caves and connected void-like areas, a significantly different environment as compared to the previous one, comprised of 10-15 meters wide caves. In the Figure we highlight the vertical shaft that the UAV must pass through to enter the rest of the cave and ERRTs efficient navigation through it. The 3D LiDAR has limited vertical field-of-view which makes completely vertical structures very hard to explore, and is why the ERRT has to guide it in a corkscrew pattern up the shaft. The Figure also highlights the 3D nature of both the environment and the total exploration path. Figure \ref{fig:cave_snaps} shows critical instances of trajectory generation throughout the mission, here highlighting examples of highly informative trajectories that lead the UAV deep into unknown areas and where calculating information-gain along trajectories has a significant impact.
ERRT selects trajectories with efficient information-gain maximizing behavior and 3D exploration in the wide interconnected cave areas, while maintaining safe navigation, efficiently moving from cave to cave without significant unnecessary maneuvering.To highlight the fast re-computation in ERRT and the smooth transition between trajectories, the Figure \ref{fig:velocity} shows a histogram of velocities during the Cave World mission. 
The Figure shows that the vast majority of instances are spent around the desired navigation speed of $\unit[0.8-1]{m/s}$ with no significant spike at the lower velocities that would indicate idling, inefficient transitioning between trajectories, or aggressive backtracking maneuvers.
More attention to the Cave World exploration of wide and interconnected voids can be found in the comparisons in Section \ref{sec:comparison}.

\subsection{Urban Exploration}
To further differentiate the evaluation environment to showcase ERRT applied in varied scenarios, we also deploy the algorithm in simulated urban environments. Of interest is the ability of ERRT to explore room-and-corridor type environments that would mimic an arbitrary indoor urban environment. 

\begin{figure}[!htbp]
    \centering
\includegraphics[width=0.9\linewidth]{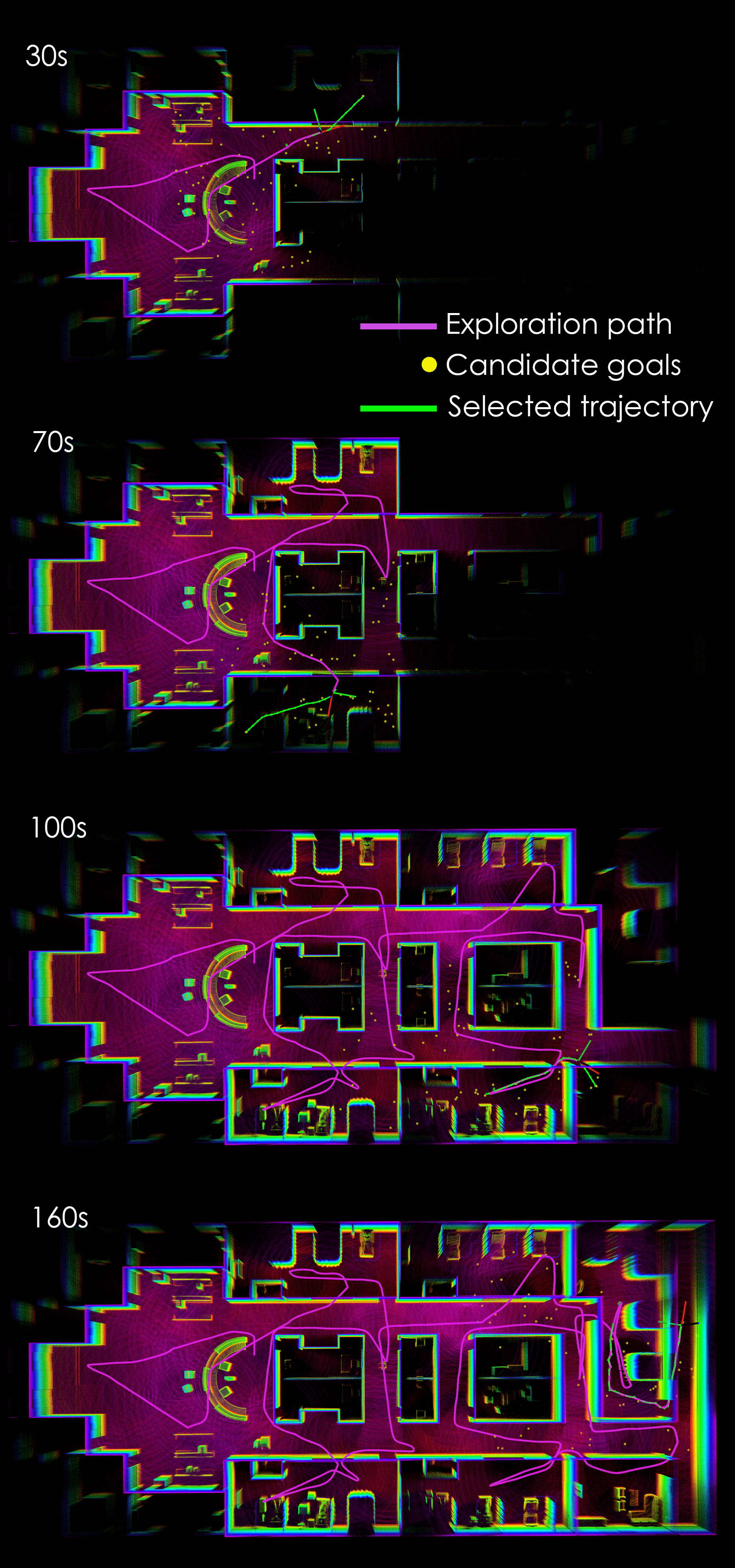}
    \caption{Timestamped Exploration run using ERRT in an urban room-and-corridor environment.}
    \label{fig:errt_hospital}
\end{figure}

\begin{figure}[!htbp]
    \centering
\includegraphics[width=\linewidth]{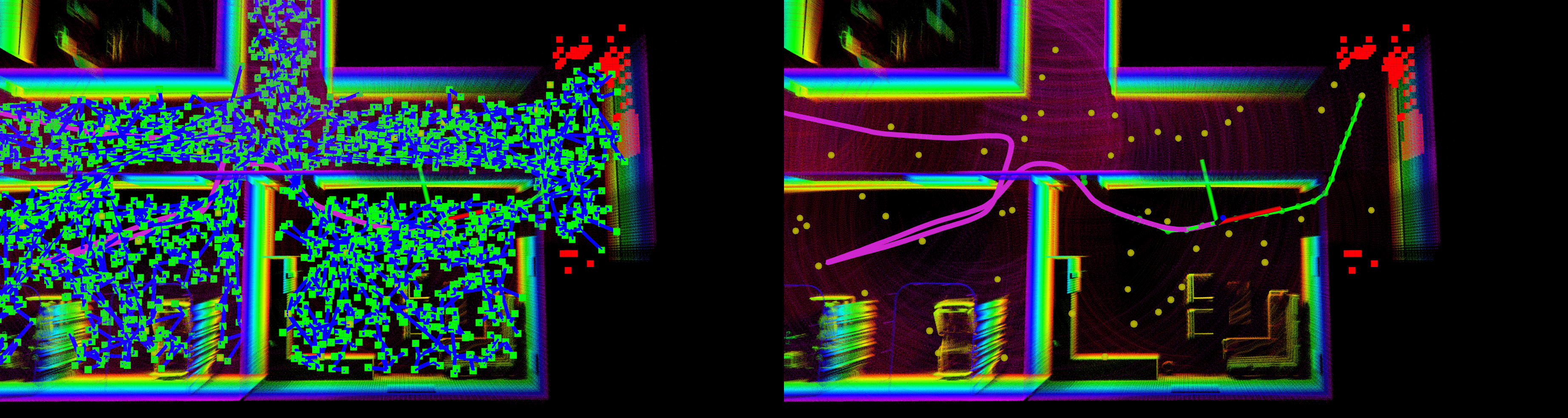}
    \caption{Snapshot of tree generation (left) and branch selection (right) with the visualization of predicted information gain (red) in the urban Hospital World.}
    \label{fig:hospital_snaps}
\end{figure}

We selected a gazebo Hospital World with multiple rooms connected by narrow doors and corridors. The ERRT simulated mission progress with set time stamps can be found in Figure \ref{fig:errt_hospital}. Here, ERRT efficiently explores and enters all rooms in the area without significant overlap and backtracking - except to greedily maximize information gain by prioritizing high-information areas first. A highlight is that ERRT does not linger in explored rooms, but efficiently moves from room to room, entering all the small rooms in the environment. We visualize one instance of trajectory generation in Figure \ref{fig:hospital_snaps} of ERRT navigating through the narrow doors and corridors, where we also visualize the predicted information gain the selected trajectories. 

\begin{figure}[!htbp]
    \centering
\includegraphics[width=0.9\linewidth]{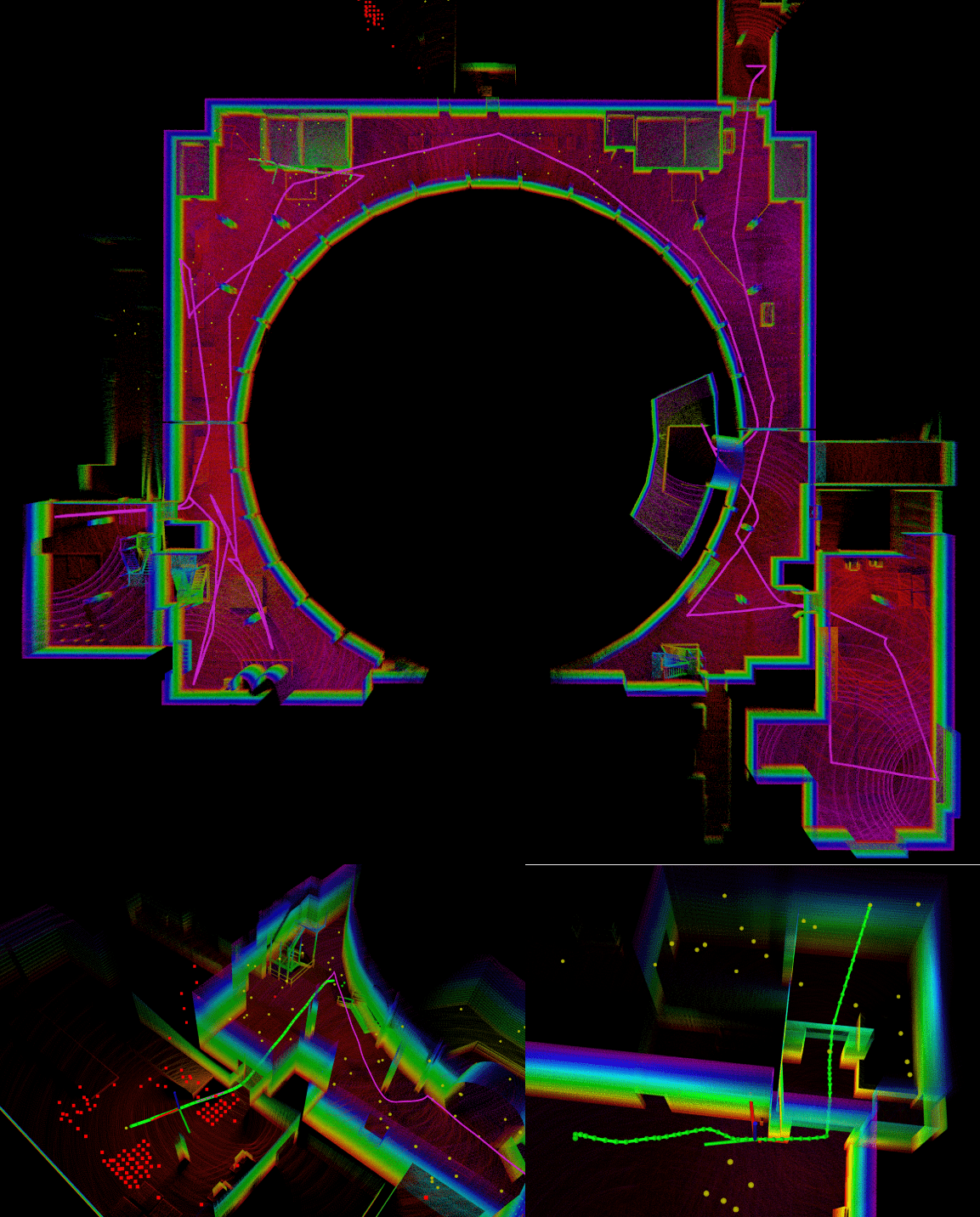}
    \caption{ERRT exploration in an environment with large 3D voids combined with very narrow constrained entrances to other rooms (top), and examples of safe trajectories through narrow entrances to new areas (bottom).}
    \label{fig:reactor}
\end{figure}

Finally, for clarity and thoroughness we want to highlight a type of environment where ERRT and similar exploration-planning strategies can struggle as for example the DARPA 2nd Stage World of an abandoned nuclear reactor seen in Figure \ref{fig:reactor}, that has large 3D voids connected by very narrow doors (around $0.8m$ openings between rooms). 
Here, driving the exploration only through information gain is sometimes not sufficient for from what a human perspective would be efficient exploration. There is simply little incentive to enter the next room without first fully exploring/mapping the much more information rich open areas as the LiDAR visibility is almost zero through the narrow entrances, and as such $\nu(\bm{\chi})$ will also be very low. Despite this difficulty, ERRT manages to both generate the exploration behavior that drives exploration into the narrow entrances and generate safe trajectories through them, in combination with efficient exploration of the large void, further highlighting ERRT deployment in different types of environments.

\subsection{Comparison}\label{sec:comparison}
We perform simulation comparisons with two other state-of-the-art exploration framework. First, the Graph-Based Planner ~\cite{dang2020graph}, developed by the winning team in the DARPA SubT Challenge, hearby denoted as GBP.  GBP is a similar sampling-based exploration-planning framework but utilizing a graph-building approach. For completeness we also include comparisons with the frontier-based strategy Rapid Exploration Framework (REF) that combines local frontier exploration with global repositioning using a risk-aware gridsearch path planner~\cite{patel2023ref}. Something worth highlighting is that both of these comparison methods use a local-global approach where a much simplified local exploration strategy is applied as long as frontiers of good quality exist in front of or around the robot, and only when local frontiers are exhausted is the more computationally heavy global strategy applied. ERRT on the other hand is only using one very general planning strategy that is local, but local on a much larger scale.

We focus the comparison on two simulated worlds - the urban Hospital World and the large subterranean Cave World to highlight exploration in varying environments. A video compilation of the exploration runs using all three frameworks in both environments can be found at \url{https://youtu.be/hmM75ohNB1Y}. The comparison metric will be simple: which framework can increase the known space the most for a set mission duration. We measure this by comparing the explored volume over time. All frameworks are deployed with a maximum exploration velocity of \unit[1]{m/s} set by the trajectory tracking controller.

\begin{figure}[!htbp]
    \centering
\includegraphics[width=\linewidth]{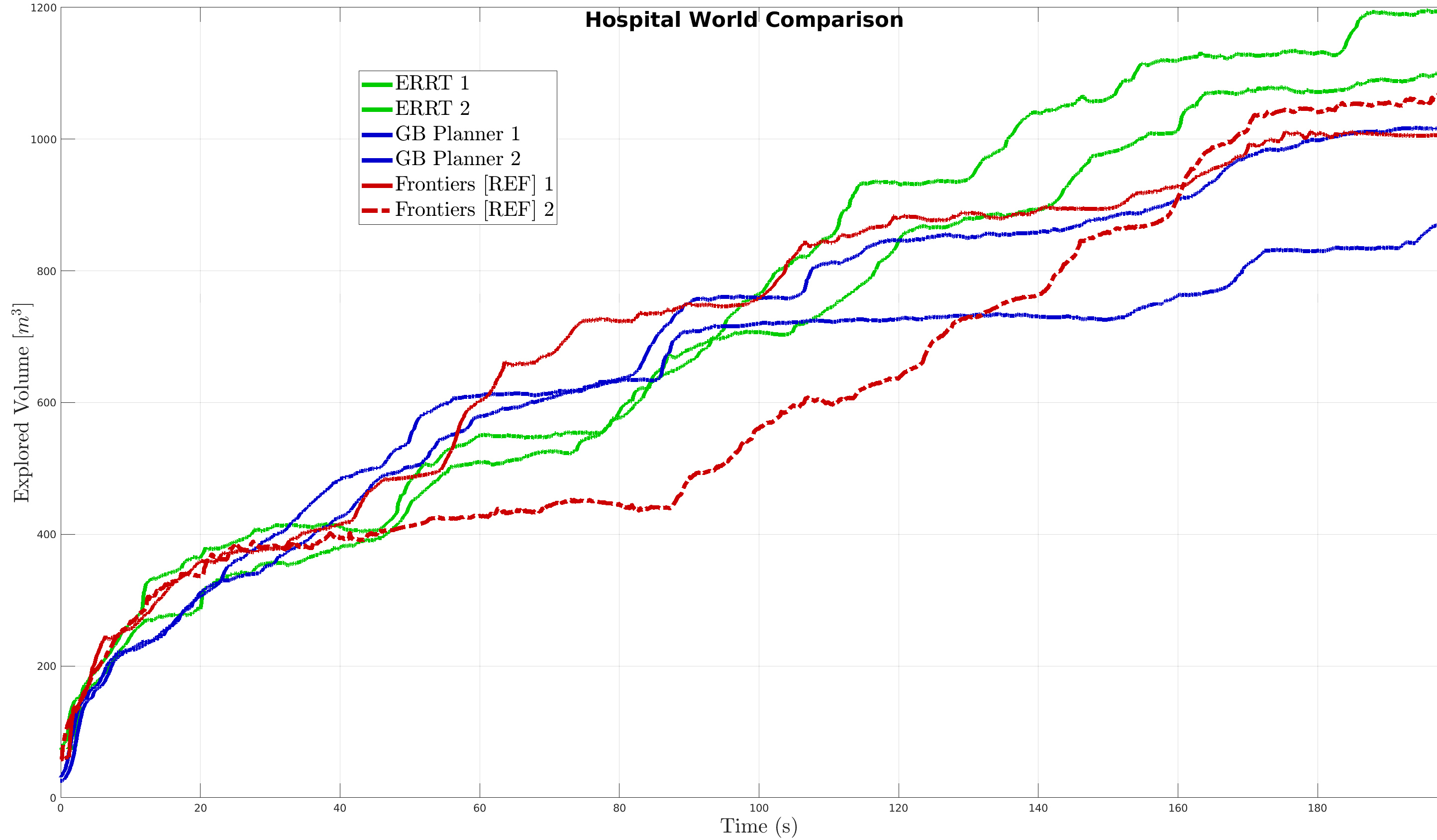}
    \caption{Information gain over time in 200s missions, in the urban Hospital World for all three frameworks. ERRT (green), GBP (blue), and REF (red).}
    \label{fig:hospital_comparison}
\end{figure}

\begin{figure}[!htbp]
    \centering
\includegraphics[width=0.9\linewidth]{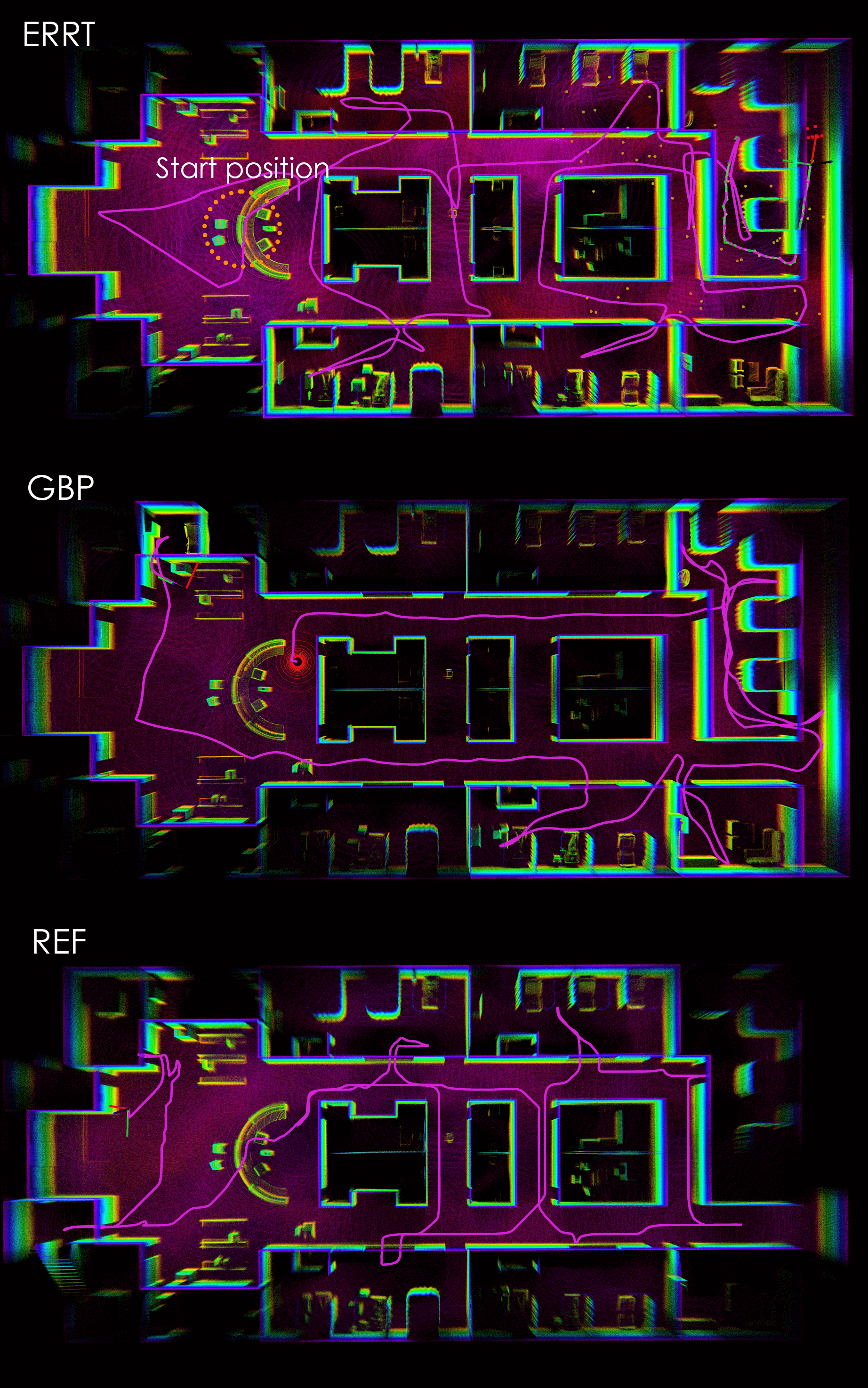}
    \caption{The state of the explored map in the urban scenario after approximately 200s for each of the three frameworks.}
    \label{fig:hospital_runs}
\end{figure}

The urban exploration comparison metric of the explored volume over time can be seen in Figure \ref{fig:hospital_comparison} where indices \textit{1,2,3} denote separate missions for each framework. Additionally, Figure \ref{fig:hospital_runs} shows the exploration trajectories and generated maps from one mission of each framework. 
The time of the exploration mission was selected to be 200 seconds based on the size of the environment. In the comparison, ERRT was around 12\% more efficient than REF and around 22\% more efficient than GBP. 
In this environment, REF shows very efficient exploration behavior but is limited due to the inability, or preference not to, enter some of the smaller narrower rooms and quickly runs out of high-information areas. GBP on the other hand spends more time than ERRT fully exploring one of the larger rooms (ex. to the right in Figure \ref{fig:hospital_runs}) by moving back-and-forth which is visualized by the lower slope on information gain for some moments in GBP. ERRT instead tries to greedily explore the most information-rich areas first - and does have more overlap in its exploration path than the other methods due to that - while GBP momentarily stays "too long" in local exploration mode looking to complete the local area before moving on. Interestingly, the two comparison methods look like they are performing more reasonable exploration behavior (following the corridors etc.), but in the end it turns out to be less efficient.

The second comparison was performed in the wide interconnected caves and voids. Here, we initially ran all three frameworks from the same starting location as for ERRT in Section \ref{sec:sim_exploration} where the UAV must enter into a vertical shaft to reach the rest of the cave.

\begin{figure}[!htbp]
    \centering
\includegraphics[width=\linewidth]{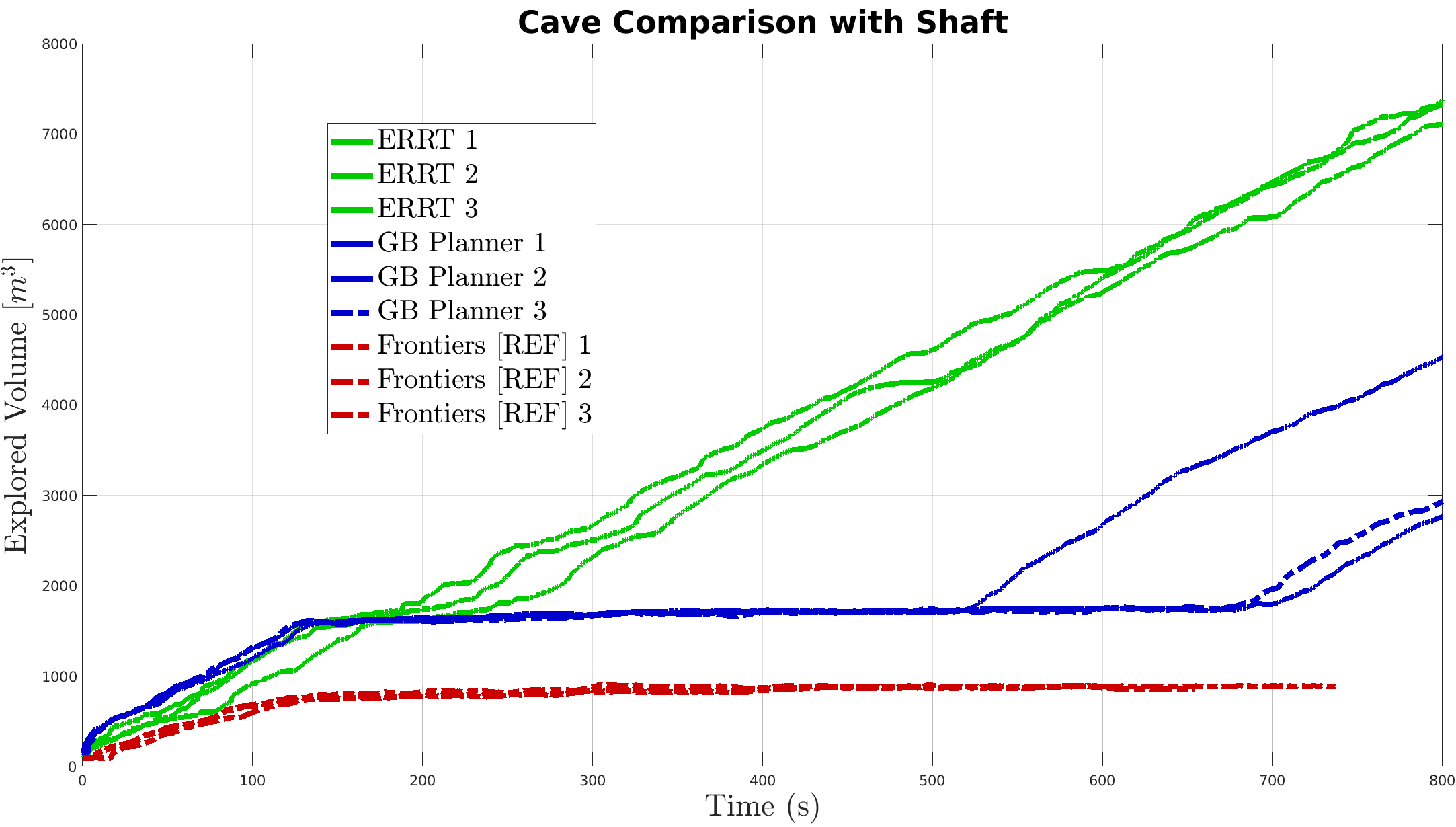}
    \caption{Information gain over time in the Cave World - starting location requires the navigation through a vertical shaft.}
    \label{fig:cave_comp_bad}
\end{figure}

\begin{figure}[!htbp]
    \centering
\includegraphics[width=\linewidth]{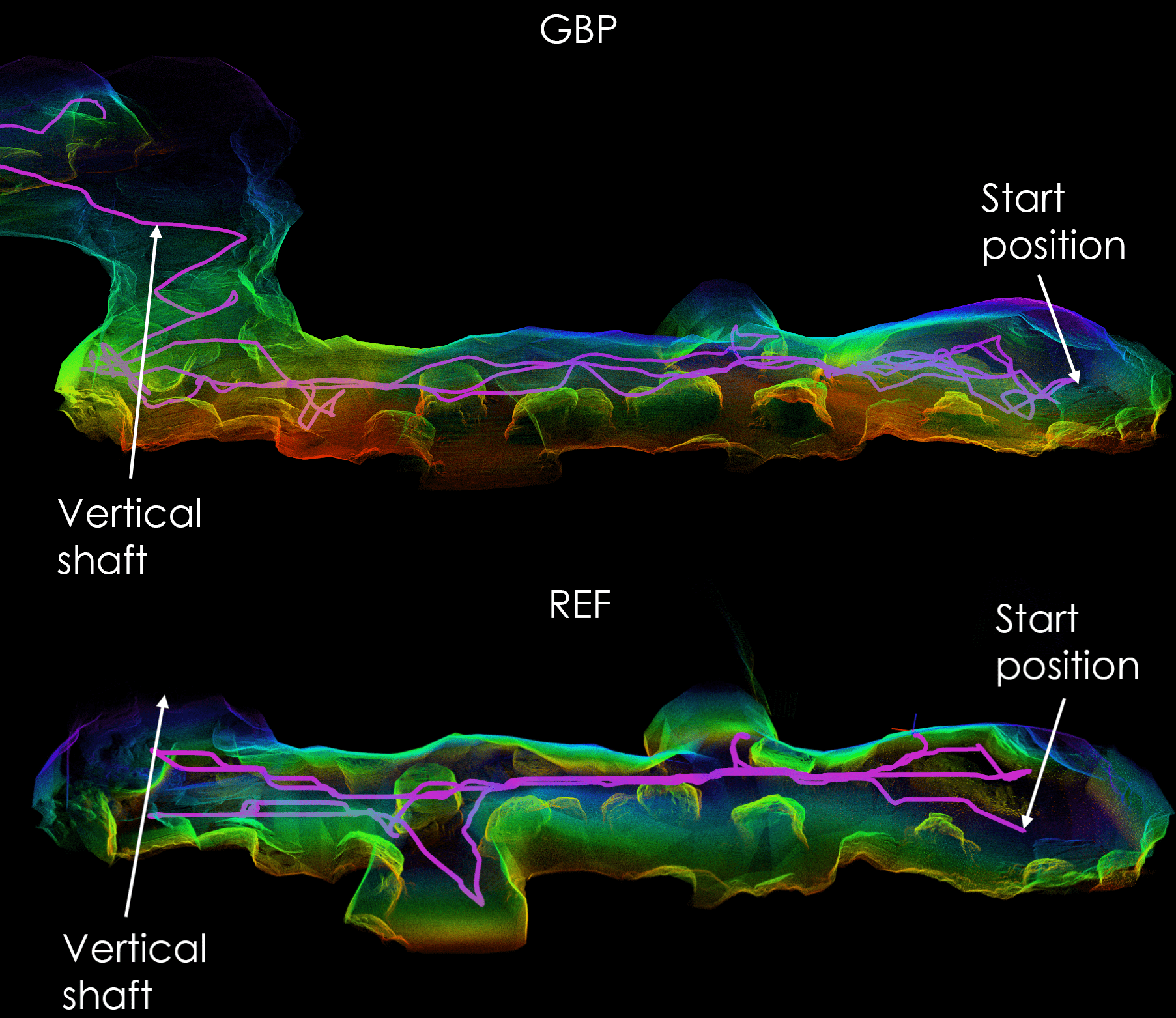}
    \caption{GBP and REF stuck on the bottom level of the cave. GBP eventually explores the shaft.}
    \label{fig:shaft_exploration_bad}
\end{figure}

Unfortunately, the comparison frameworks would get stuck exploring the first level for a very long time trying to achieve complete coverage before entering the shaft, which is visualized in Figure \ref{fig:cave_comp_bad} for the explored volume, and Figure \ref{fig:shaft_exploration_bad} showing GBP (stuck for a long time) and REF (never enters the shaft) exploring the first level, to be compared to ERRT efficiently exploring the 3D area and the shaft in Figure \ref{fig:cave_world} with no backtracking. We can see in Figure \ref{fig:cave_comp_bad} that the ERRT exploration barely slows down as a consequence of the vertical structure.
Due to the 3D LiDAR field-of-view, the vertical shaft is a very difficult as frontiers generated near the shaft will have limited numbers of unknown voxels inside the narrow horizontal field-of-view of the LiDAR and thus low information gain. As ERRT evaluates information gain along trajectories, more unknown voxels can be seen from different angles for a selected exploration trajectory promoting the UAV to enter into the vertical shaft.

\begin{figure}[!htbp]
    \centering
\includegraphics[width=\linewidth]{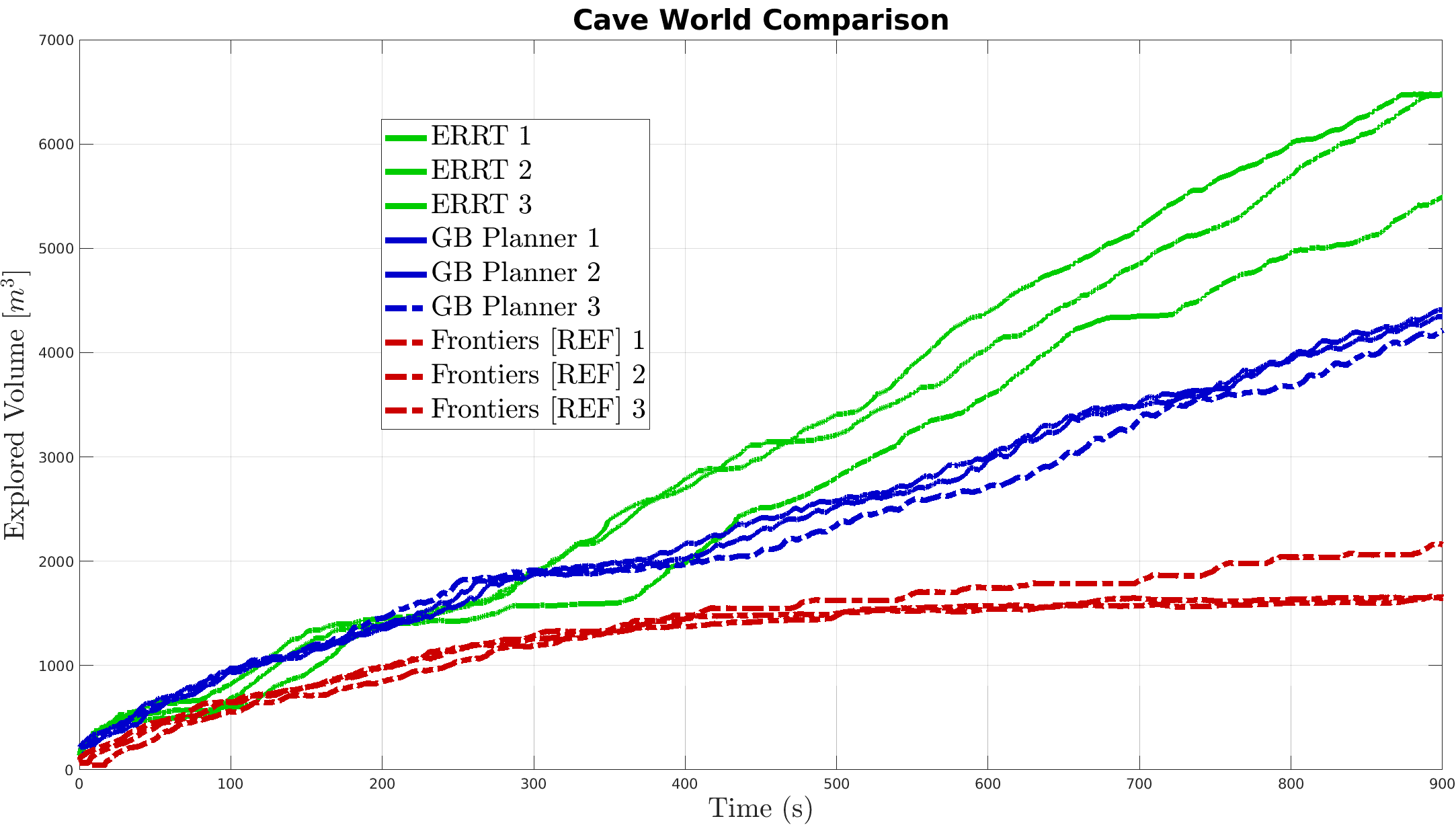}
    \caption{Information gain over time in 15 minute missions, in the large-scale Cave World for all three frameworks. ERRT (green), GBP (blue), and REF (red).}
    \label{fig:cave_comparison}
\end{figure}

As we still wanted to properly evaluate the competing methods in the wide caves, we blocked off the vertical shafts in the cave world and re-started the missions on the other side of the shaft. The comparison metric of explored volume over time can be found for all runs in Figure \ref{fig:cave_comparison}. In general, both REF and GBP will stay in local exploration mode at inappropriate moments which reduces their total information gain, or not fully utilizing the 3D nature of the environment to maximize the explored volume. REFs global grid-search path planner also starts to have significant computation time issues as the size of the mission expands past a few hundred meters. Overall though, GBP showcased very good exploration behavior but mainly explored the voids in a single plane with limited 3D maneuvering that limits coverage.
In the wider caves, ERRTs next-best-trajectory methodology can really shine since much information is available and managing to maximize greedy behavior \textit{along} exploration trajectories in a relatively large 40m sampling space, more often than not selecting long trajectories that move in 3D to maximize LiDAR field-of-view coverage. This leads to ERRT outperforming the other methods with significant margins.
ERRT outperforms REF by around 240\% in exploraton efficiency, and the GBP by around 50\%.

Fair comparisons in realistic and complex environments are hard to do, and the tuning of the frameworks has a large impact. These numbers should mainly highlight and validate the general ERRT concept for large-scale exploration in complex 3D environments as competitive to the state-of-the-art algorithms, and when the environment is right, can also significantly outperform them.

\section{Experiment Results\label{sec:experiments}}
\subsection{Hardware and Complete Autonomy Architecture\label{sec:setup}}
\begin{figure}[!htbp]
    \centering
\includegraphics[width=\linewidth]{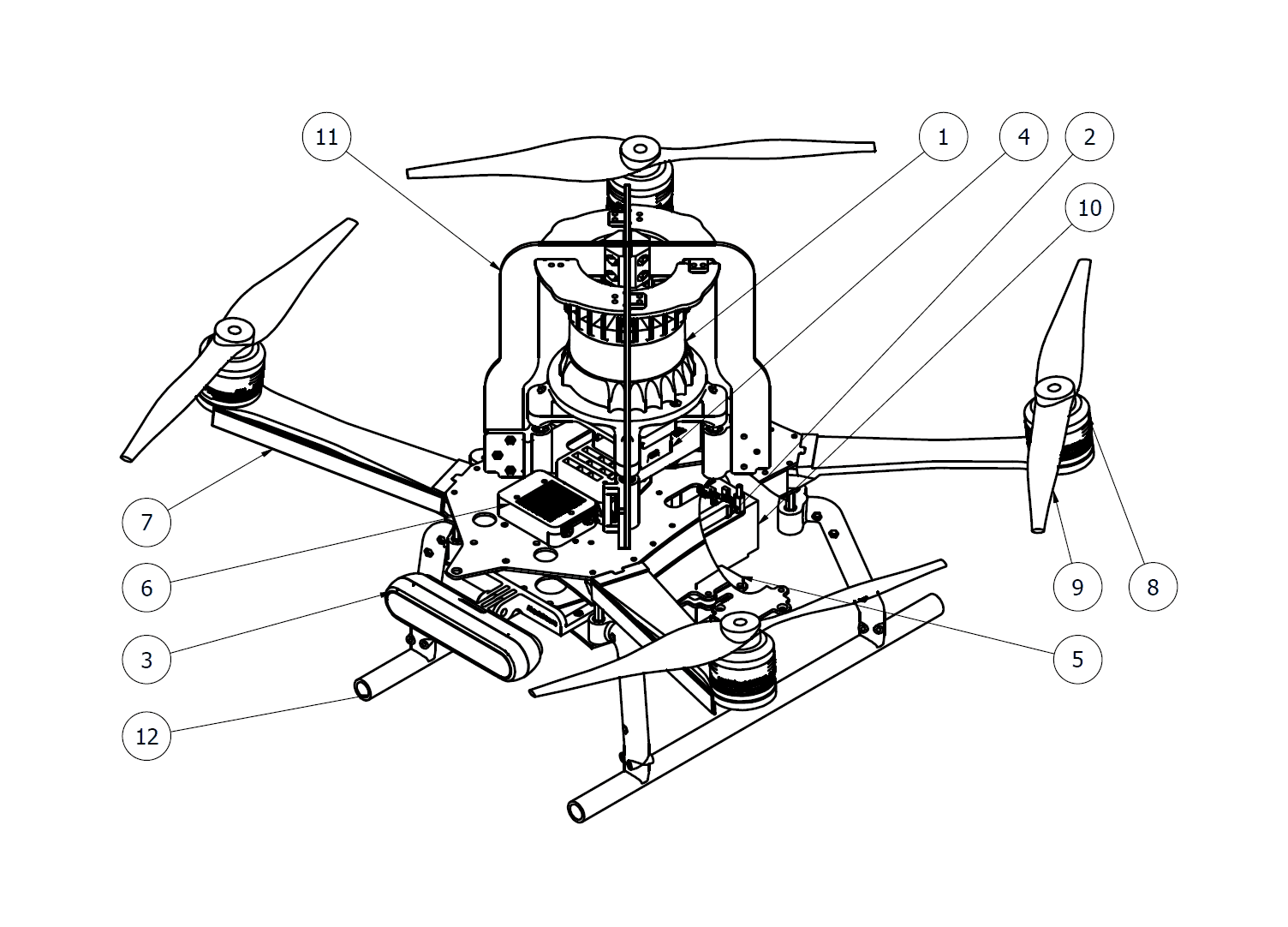}
    \caption{The custom-built UAV platform used for experimental validation. 1 - Ouster OS1 3D LiDAR, 2 - Intel NUC, 3 - Intel Realsense D455, 4 - Pixhawk Cube Flight Controller, 5 - Garmin singlebeam LiDAR, 6- Telemetry module, 7 - LED strips, 8- T-motor MN3508 kV700, 9- 12.5in Propellers, 10 - Battery, 11  - Roll cage, 12- Landing gear}
    \label{fig:shafter}
\end{figure}

\begin{figure}[!htbp]
    \centering
\includegraphics[width=\linewidth]{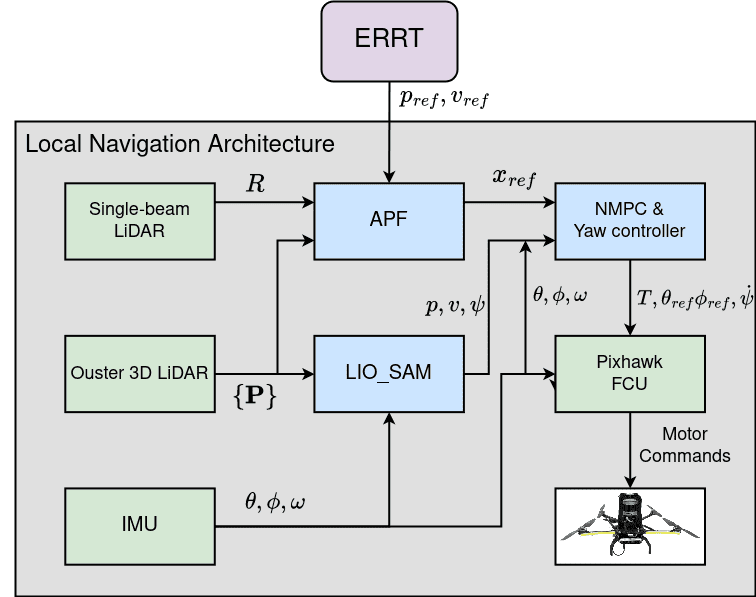}
    \caption{Local hardware, state-estimation, and navigation framework that ERRT is added on top of. The complete local autonomy framework enables experimental field evaluations.}
    \label{fig:local_arch}
\end{figure}

The custom built UAV platform that was used for hardware experiments can be seen in Figure \ref{fig:shafter}. The relevant hardware components are an Ouster OS1-32 3D LiDAR with 45$^\circ$ vertical field of view, an Intel NUC 10 BXNUC10I5FNKPA onboard computer with an Intel i5 1.6Gz processor, and a Pixhawk\cite{meier2011pixhawk} Cube FCU with internal IMU running PX4\cite{meier2015px4}. The UAV has a size radius of maximally $\unit[0.38]{m}$ (diagonally), but to ensure safe navigation we define the $V_\mathrm{safe}$ with a robot radius of $r_\mathrm{robot} = \unit[0.6]{m}$ implying that the center of the UAV should be at least $\unit[0.6]{m}$ from any occupied or unknown voxels. During real-world experimentation, we prioritize the safety of the robot, and as it is operating under real localization and control errors, the safety distance is increased to compensate.
The full supportive autonomy architecture can be seen in Figure \ref{fig:local_arch} and is fundamentally the same as described in \cite{lindqvist2022adaptive}. Position and velocity references $p_\mathrm{ref}, v_\mathrm{ref}$ are sent to an Artificial Potential Field (APF) (as the attractive force) that uses raw LiDAR data to generate repulsive forces and acts as an additional safety layer in case of momentarily poor voxel mapping. The APF generates the collision-free state reference $x_\mathrm{ref}$ to a Nonlinear MPC~\cite{lindqvist2020dynamic} (fundamentally formulated the same way as the actuation-trajectory NMPC in Section \ref{sec:actuation}) which in turn communicates with the FCU (Flight Control Unit) at $\unit[20]{hz}$ through thrust, pitch references, roll references and yawrate commands as $u = [T, \phi_\mathrm{ref}, \theta_\mathrm{ref}, \Dot{\psi}]$.
The robot state $\hat{x} = [p,v,\phi, \theta, \psi, \omega]$, consisting of position, velocity, Euler angles, and angular rate states is estimated by the state-of-the-art LiDAR-Intertial Odometry framework LIO-SAM~\cite{shan2020lio} and the onboard IMU (Inertial Measurement Unit). 

The ERRT tuning used in experiments was set up to enable the program to run on more limited hardware, and can be found in Table \ref{table:exp_param}. It was tuned to have a computation time around 0.8-1 seconds.

\begin{table}[!ht]
\begin{center}
\caption{Critical ERRT parameters used for field experiments.}
 \label{table:exp_param}
\begin{tabular}{ |c|c|c|c|c|c|c|c| } 

 $V^l_\mathrm{map}$ & $n_\mathrm{traj}$ &  $S_r$ &$size(\bm{N})$ & $d_\mathrm{info}$ & $K_d$ & $K_i$ & $K_u$ \\ 
 24m & 40 & 8 & 1000 & 6m & 0.3 & 0.8 & 0.1 \\ 
\end{tabular}
\end{center}
\end{table}

\subsection{Field Evaluation\label{sec:field_evaluation}}
\begin{figure}[!htbp]
    \centering
\includegraphics[width=\linewidth]{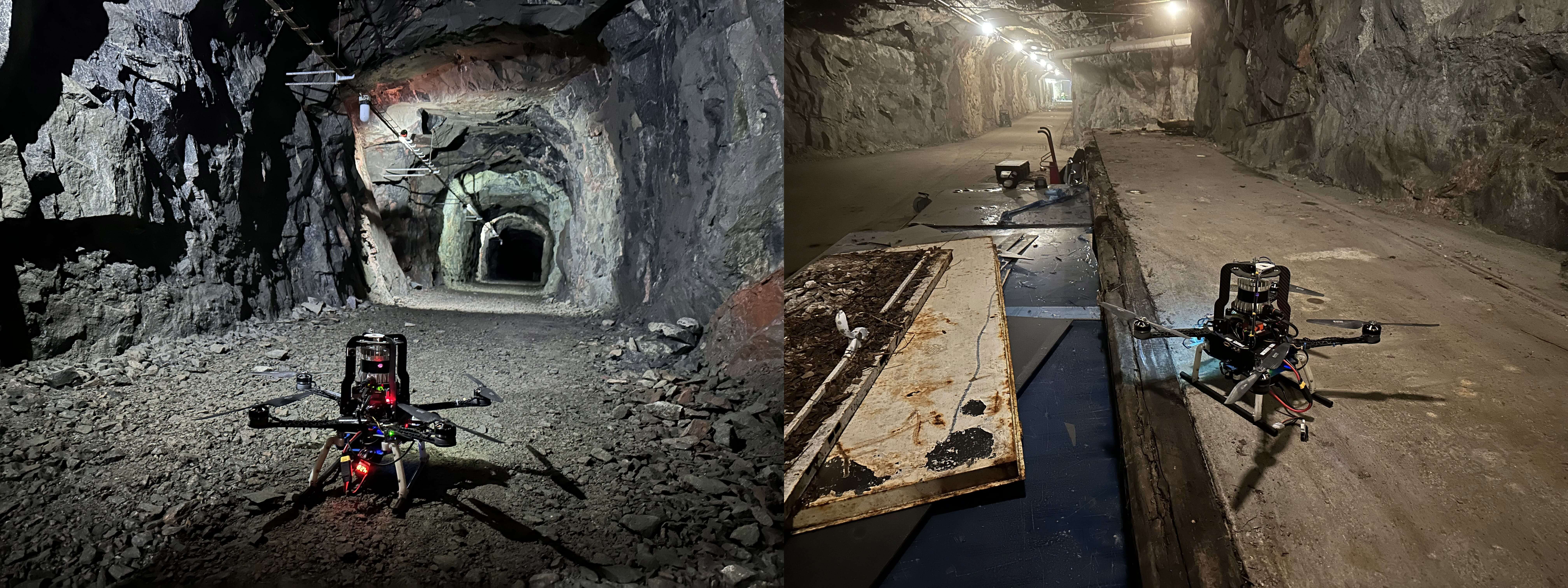}
    \caption{Field Evaluation Environments for the two performed experiments. Narrow constrained tunnel (left), open but cluttered area with connected tunnels (right).}
    \label{fig:mj}
\end{figure}

Hardware experiments were performed in a relevant subterranean and GPS-denied area at Mjölkuddsberget, Luleå, Sweden, a sample of which can be seen in Figure \ref{fig:mj}. Two experiments were performed, where ERRT is deployed on the UAV platform seen in Figure \ref{fig:shafter} and supported by the autonomy stack in Figure \ref{fig:local_arch}. 

\begin{figure}[!htbp]
    \centering
\includegraphics[width=\linewidth]{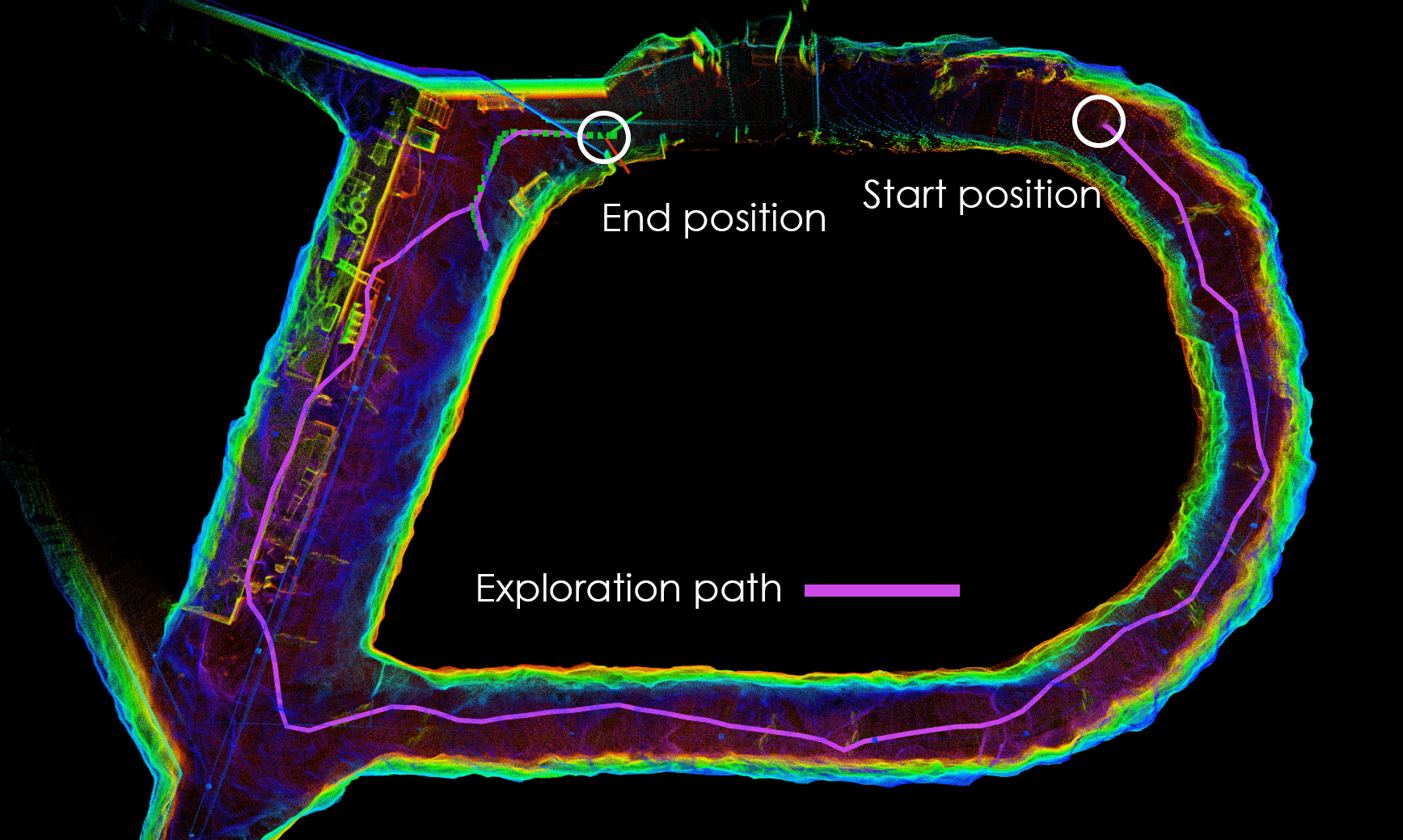}
    \caption{Exploration path from local exploration in the field, in curving tunnel area with a larger void. Exploration path length is around $\unit[160]{m}$.}
    \label{fig:field_rondell_full}
\end{figure}

\begin{figure}[!htbp]
    \centering
\includegraphics[width=0.8\linewidth]{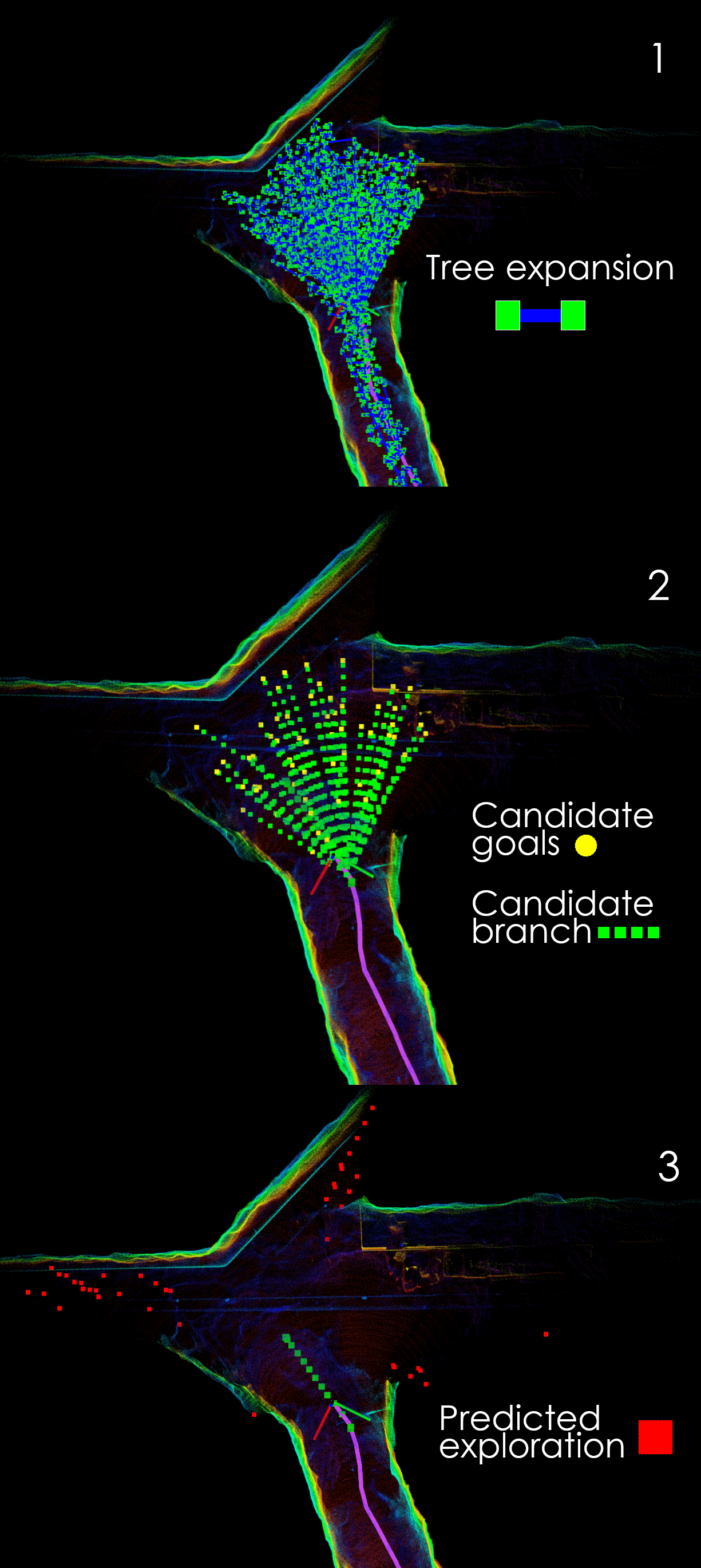}
    \caption{The ERRT stages in a field environment with a junction. 1) Tree expansion, 2) candidate goals and candidate trajectories, and 3) selected exploration trajectory with expected explored volume (red).}
    \label{fig:field_rondell_stages}
\end{figure}

\begin{figure}[!htbp]
    \centering
\includegraphics[width=\linewidth]{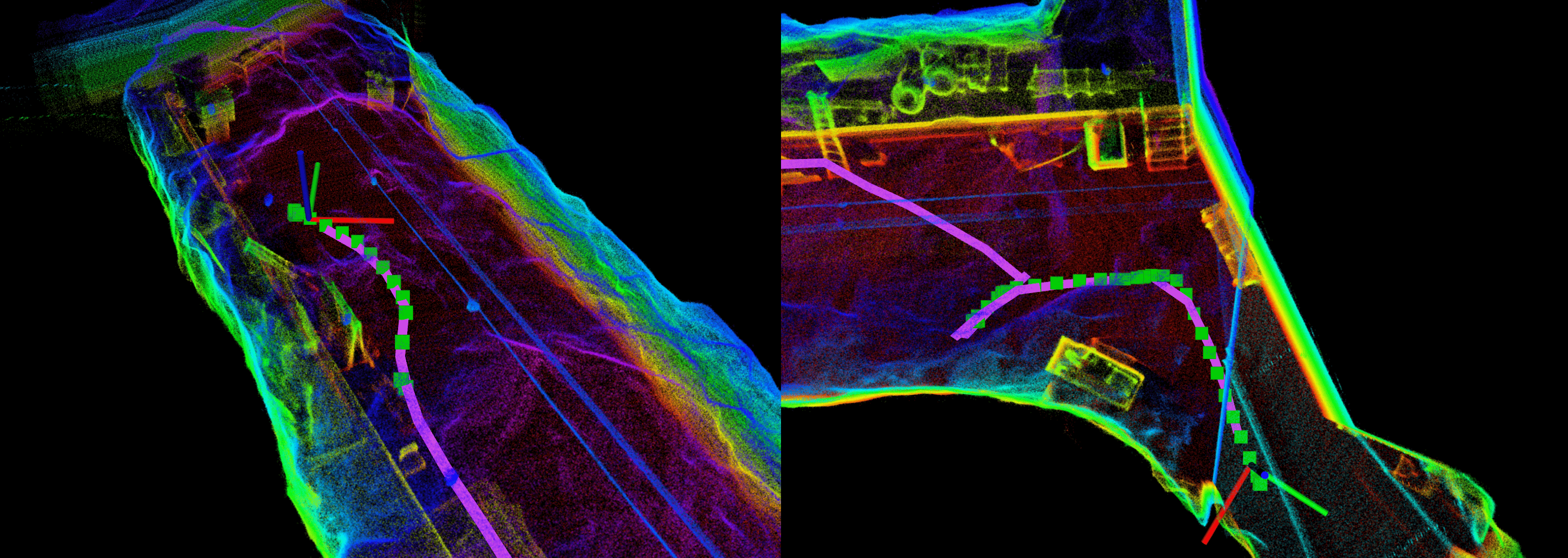}
    \caption{ERRT dynamic and robot-safe trajectory generation around small/complex obstacles in obstacle-rich and narrow environments in the field.}
    \label{fig:field_rondell_obstacles}
\end{figure}

The first experiment was performed in an area of a curving tunnel environment that also includes a larger void area with various obstacles, and the second experiment was performed in a very narrow, straight, and constrained mining tunnel that includes also a junction area. Figure \ref{fig:field_rondell_full} shows the exploration run in the curving tunnel area where ERRT successfully provided navigation through the tunnel with no instants of backtracking, efficiently avoiding obstacles, and maximizing information gain highlighted by paths generated instantly deep into the open void area as opposed to entering the narrower tunnels at the junction. The total exploration path length was around $\unit[160]{m}$. Figure \ref{fig:field_rondell_stages} again demonstrated the ERRT concept here in a real field deployment scenario; showing the 3D tree expansion of robot-safe branches, the extracted and improved trajectories that terminate at sampled goals, and the correctly selected information-gain maximizing trajectory that is also relatively short in distance and low in actuation, into the wider junction area. Additionally, Figure \ref{fig:field_rondell_obstacles} shows two instants of ERRT navigating around obstacles in the cluttered environment, providing both robot-safe and dynamical  (from the NMPC module) paths. The average computation time for ERRT in this mission was approximately tuned to be $\unit[1]{s}$.

\begin{figure}[!htbp]
    \centering
\includegraphics[width=\linewidth]{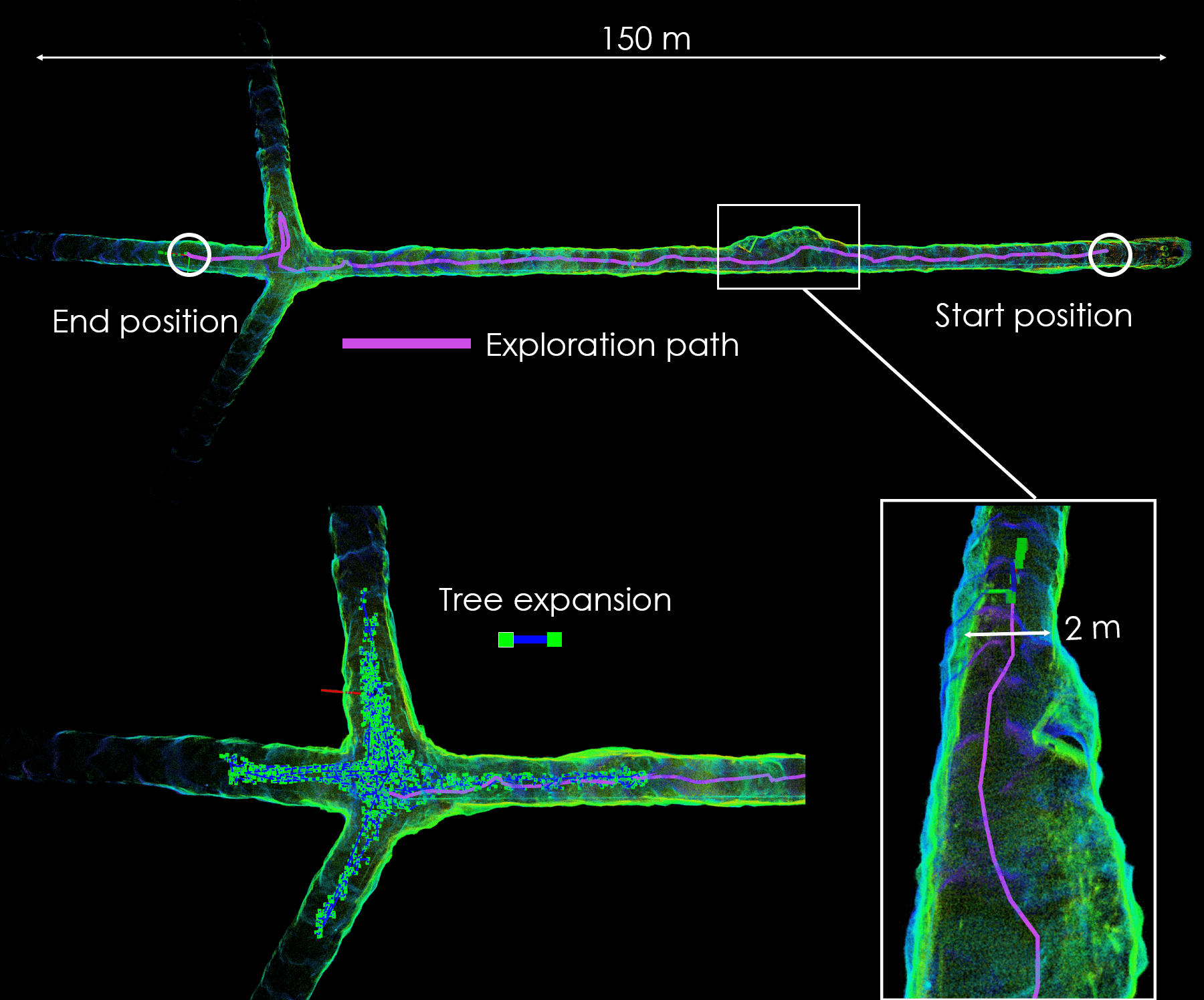}
    \caption{ERRT exploration path in a very constrained and narrow field environment with a total exploration path of $\unit[170]{m}$ (top), and samples of tree expansion in the juction and path selection in an extra narrow section during the exploration run (bottom)}
    \label{fig:field_narrow_all}
\end{figure}

Figure \ref{fig:field_narrow_all} shows the generated map from exploration in a very narrow and constrained tunnel area around $\unit[2]{m}$ wide with various obstructions, and critical snapshots during exploration of tree expansion and safe navigation. In this run ERRT also safely navigates through the tunnel for a total exploration path length of around $\unit[170]{m}$ with a continued forward exploration without unnecessary side-to-side movements or backtracking, while keeping a safe distance from any walls and keeping in the middle of the tunnel. This provides and example of that the ERRT safe navigation behavior in very narrow and constrained areas demonstrated in simulation is consistent with results from the field. Only at one moment in the junction area did ERRT generate backtracking behavior (going from one junction branch to the other), but in this case this is warranted as to first greedily explore the wider junction area as opposed to the narrow tunnel.

\section{Performance, Tuning, and Computation}
This section will aim to give the reader some insights into how various tuning parameters in ERRT has an effect on the exploration performance. These parameters include the number of nodes in the RRT, which we can denote as \texttt{size($\bm{N}$}), the number of candidate goals/trajectories $n_\mathrm{traj}$, the size of the local sampling space $V^l_\mathrm{map}$, and the resolution of the occupancy map voxel size $v_s$. Another key aspect when it comes to computational complexity is the \texttt{depth} of performing collision- and information gain checks. The UFOmap \cite{duberg2020ufomap} framework operates very well with high resolution (and in all the above experiments and simulations it is run at between $\unit[0.05]{m}$ and $\unit[0.1]{m}$ voxel size) but performing volumetric checks for robot-safe tree generation and the evaluation of information gain $\nu(\bm{\chi})$ along trajectories can be computationally expensive if run at a high resolution. Instead, those checks can be performed at a certain depth in the OcTree, which can greatly improve performance, especially for the information gain. 

For self-comparison, we focus on the DARPA Cave World, the simulation world seen in Figure \ref{fig:cave_world}, as the large and geometrically complex interconnected voids and caves are perfect for the ERRT "next-best-trajectory" methodology. 
In general, the behavior and resulting computation time of ERRT can be configured to fit very varying specifications when it comes to the available computational hardware, but also whether the algorithm should focus on rapid very local exploration as compared to larger sampling spaces and more candidate solutions. We can also analyse what the effect of evaluating the information gain along the trajectory at the expense of more computation has on the performance, as opposed to only evaluating at the trajectory end-points - the more common approach. We should note that these simulations are run on a powerful machine to enable real-time simulations in the huge SubT Worlds and as the such numbers for computation time can vary for the exact tuning.

\begin{figure}[!htbp]
    \centering
\includegraphics[width=0.9\linewidth]{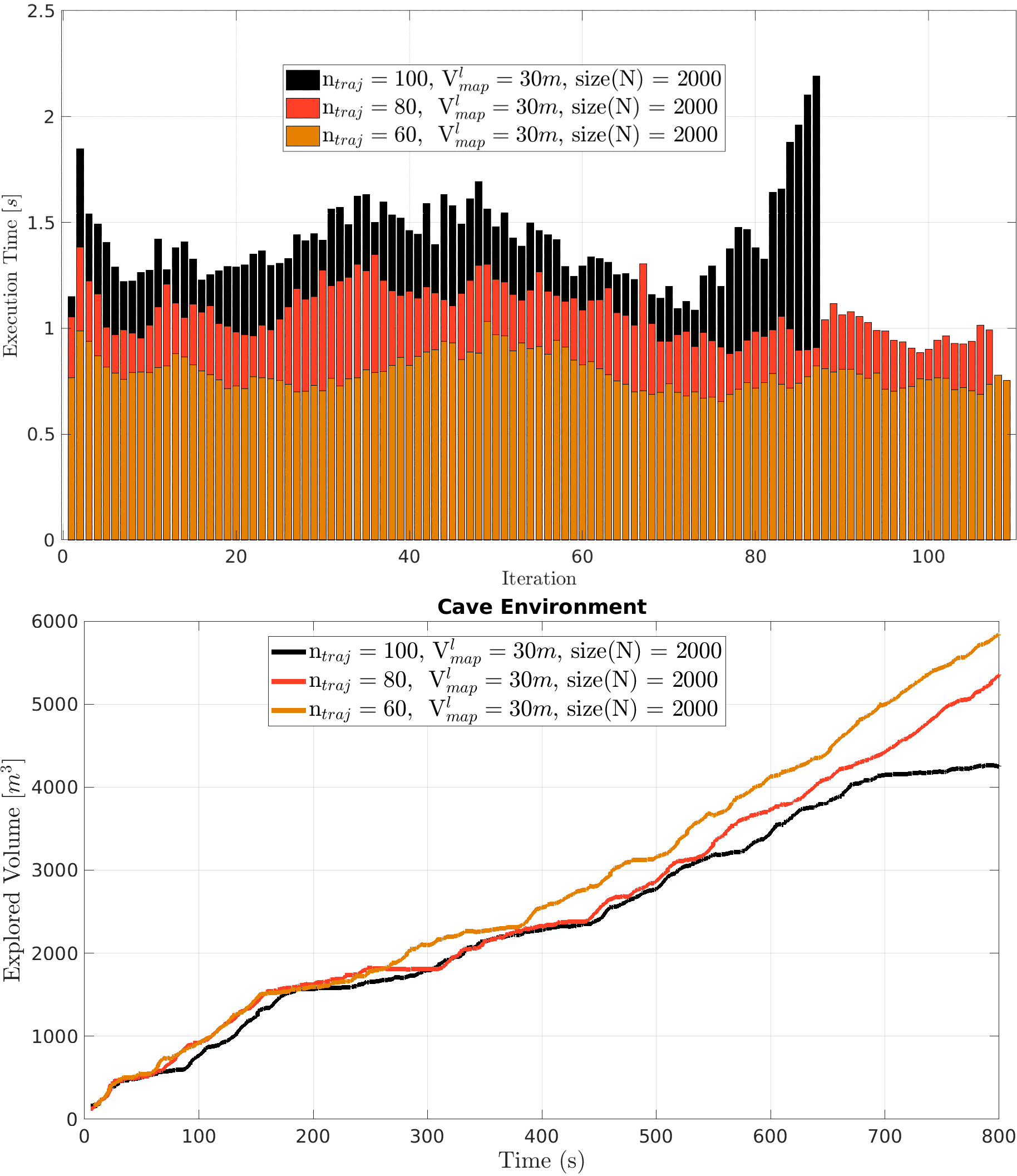}
    \caption{Comparison on computation and exploration efficiency for varying selections of $n_{traj}$ while keeping all other parameters the same.}
    \label{fig:ntraj}
\end{figure}

First, one of the critical ERRT components is the number of candidate branches $n_{traj}$ to first generate (through goal sampling) and then evaluate. Figure \ref{fig:ntraj} shows the effect on computation time and over-time information gain.

\begin{figure}[!htbp]
    \centering
\includegraphics[width=0.9\linewidth]{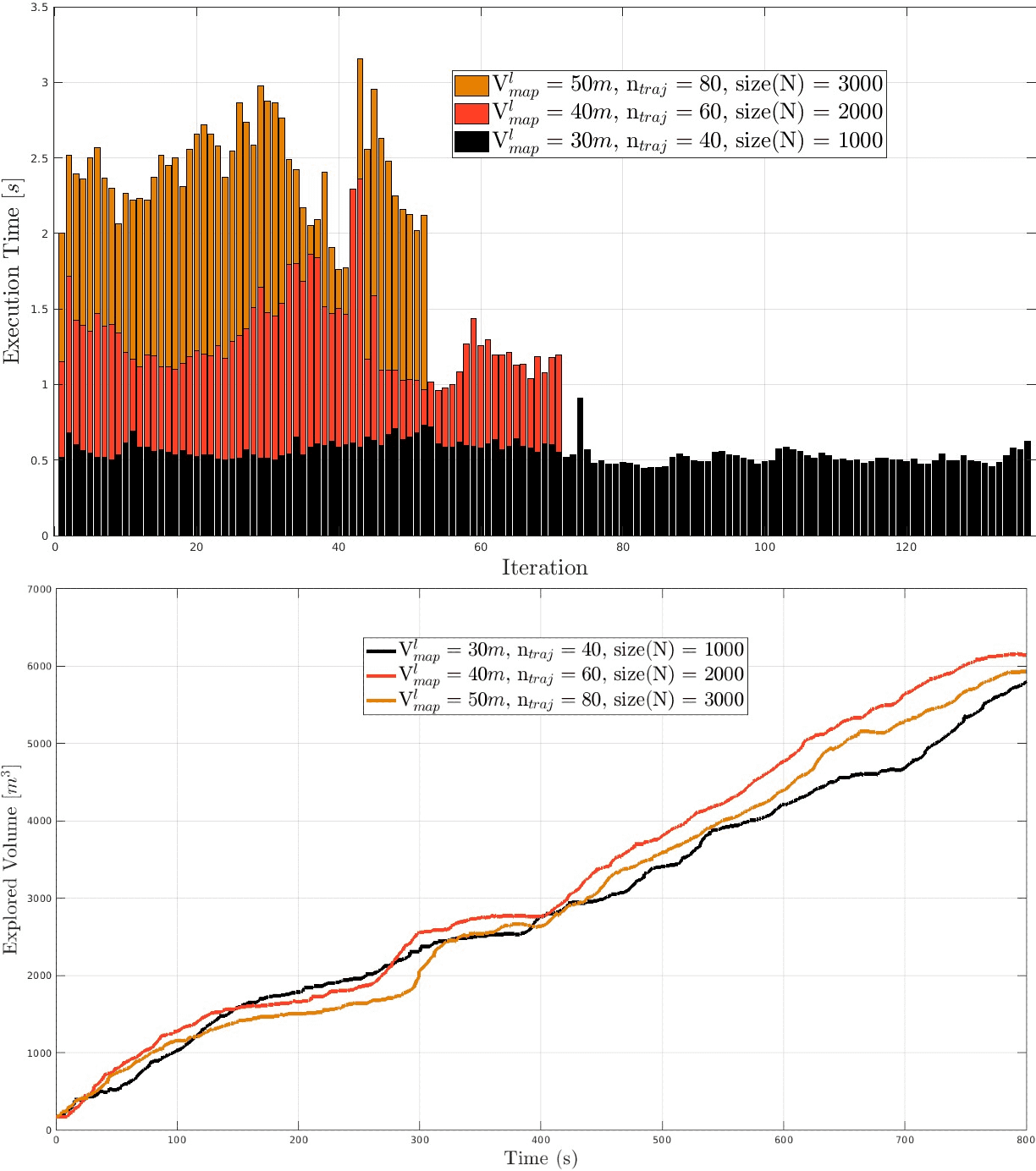}
    \caption{Comparisons for three different ERRT tunings - comparing faster computation (less idling) to better trajectory selection.}
    \label{fig:samplingspace}
\end{figure}

We can see that, as expected, the overall computation time of ERRT is highly related to the selection of $n_{traj}$ as information-gain calculations along candidate branches dominate the computation. An interesting result is that if we keep the sampling space the same size, the number of candidate solutions has a negative effect on the exploration efficiency due to the longer computation times (here a 40\% decrease in efficiency for a ~80\% longer average computation). Thus, the extra computational effort is wasted and the de-acceleration/acceleration and idling time during computation has a significant effect. 
Alternatively we can correspondingly increase the local sampling space $V^l_\mathrm{map}$ and the size of the generated tree $\bm{N}$ together with the number of candidate branches $n_\mathrm{traj}$, which is shown in Figure \ref{fig:samplingspace}. Here, despite a 400\% difference in computation time between fastest and slowest, the exploration efficiency remains close to the same. This can be attributed to a selection of better trajectories - which in general are also longer, reflected by fewer instances of re-calculation.
Here the intermediary tuning slightly wins out, which is also close the tuning used in the simulation evaluations and comparisons in Section \ref{sec:simulation}.

\begin{figure}[!htbp]
    \centering
\includegraphics[width=0.9\linewidth]{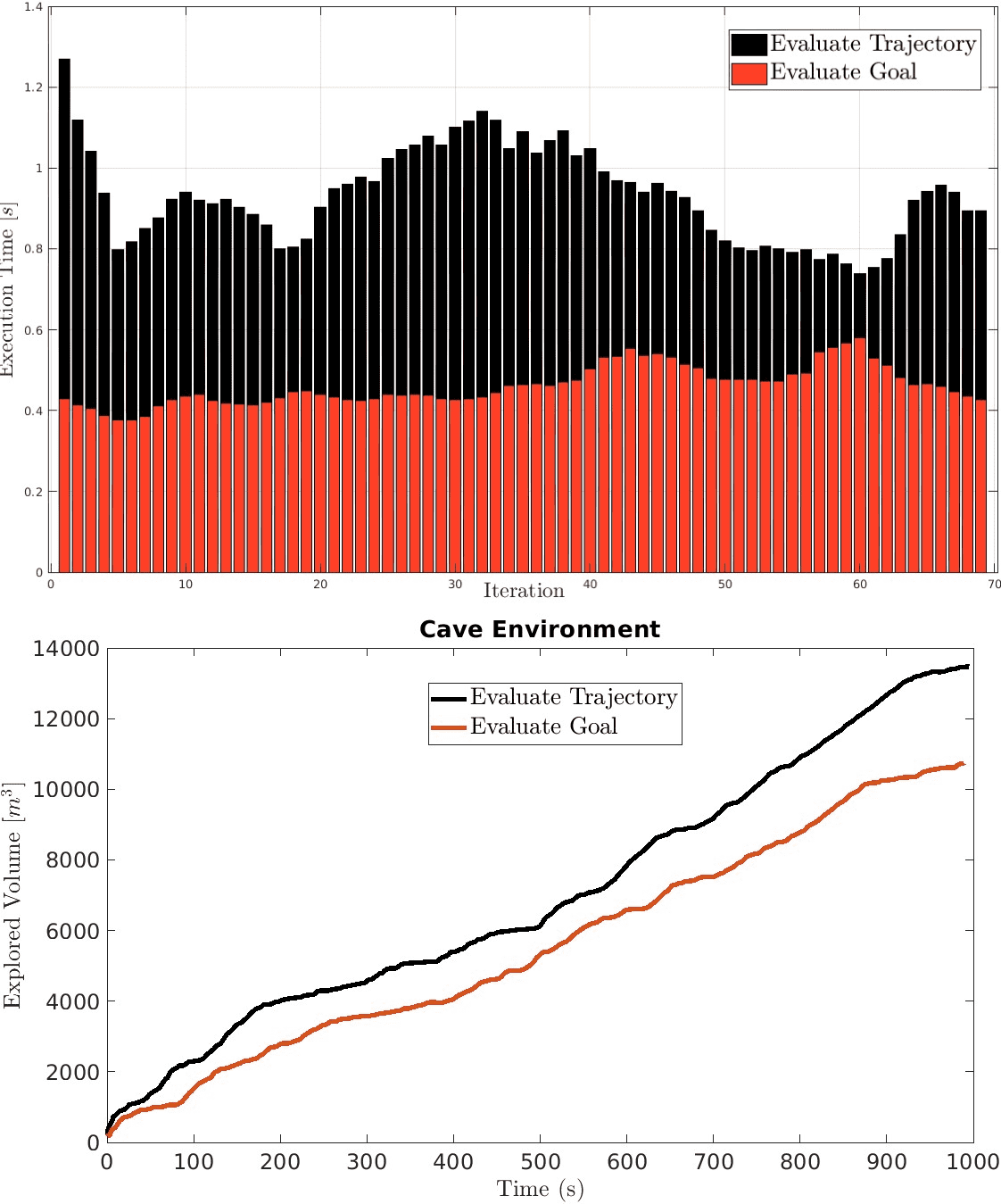}
    \caption{Comparison for evaluating information gain as $\nu(\chi)$ or only at $\nu(g^c)$. In this case, spending more computation to find better trajectories is beneficial.}
    \label{fig:along}
\end{figure}

Finally, and perhaps most importantly, we investigate how the evaluation of information gain \textit{along} the trajectories as opposed to only at the goals effect the exploration efficiency. This is visualized in Figure \ref{fig:along} where all other tuning parameters are kept the same but only the method of information gain evaluation differs. Here, an almost 100\% increase in computation time results in an over 30\% \textit{more} efficient exploration mission. 
We can thus state, based on Figure \ref{fig:samplingspace} and \ref{fig:along}, that the selection of better exploration trajectories matters as much as fast computation when it comes to time-constrained exploration missions.

\section{Limitations}\label{sec:limitations}
While the ERRT framework demonstrates efficient exploration and planning behavior in both simulations and experiments there are a set of limitations that lead to future works. For instance, while ERRT can be run on limited hardware as shown in Section \ref{sec:field_evaluation}, there were some spikes of higher execution time mainly related to performing a very high number of volumetric collision checks on the 3D occupancy map throughout the program. The computation time highly depends on the type of environment but also on the tuning of the framework. As a future work, using more complex sampling methods would be very interesting, as for example the work in \cite{wang2020neural} that trained a neural network to perform non-uniform sampling for more efficient tree expansion which could see a great application in an ERRT-like program. Additionally, the trajectory selection and as such also the performance of the framework is highly dependant on the information gain parameter $K_\mathrm{i}$  (that depends also on the voxel size as $\nu(\bm{\chi}_j)$ is related to the \textit{number} of unknown voxels in sensor range). An interesting investigation could be to reformulate the revenue/cost function \eqref{eq:realproblem} from a pure linear combination of terms to be less dependant on tuning the relative gains without losing the ability to easily tune the framework to the users needs. 
Another large limitation of ERRT relates directly to our initial problem formulation in \eqref{eq:realproblem}. ERRT fundamentally only tries to answer the Next-Best-Trajectory problem in the context of that formulation and has no consideration of any other parameters. One such studied parameter is generating trajectories that reduce localization uncertainty~\cite{carrillo2015autonomous,carrillo2018autonomous, papachristos2019localization} often by promoting motion through previously known or feature-rich areas. This links the navigation problem with the SLAM problem in an interesting way, and is a direction of future work of ERRT - perhaps to include the number of well-defined LiDAR features that will be within field-of-view of the LiDAR along all candidate branches as a reward (negative cost) in addition to the information gain. Similarly, the ERRT problem formulation has no concept of complete coverage/mapping before moving on to more information-rich areas, while there are other works that do take that into consideration~\cite{hollinger2014sampling, schmid2020efficient}. If the metric of complete exploration of a limited map area was used as opposed to exploration efficiency in limited time missions, other methods could outperform ERRT - but we highlight that this is not the main targeted application of ERRT.
Finally, ERRT in the current version has no integrated globalisation strategy in its program and as such can not continue exploration if a dead end is encountered where no junction or unexplored area can be reached within the local sampling space $V^l_\mathrm{map}$ (although this can be set relatively high if desired). This is a clear direction of future work.

\section{Conclusions} \label{sec:conclusion}
In this manuscript we have presented a fully realised Next-best-trajectory method for robot exploration: ERRT, based on a method of maximizing information gain while minimizing distance and model-based actuation along the trajectory. The method has been evaluated in simulations of complex, narrow, and large scale simulation environments, as well as in real-life experiments in relevant subterranean narrow areas that highlight the challenges of safe navigation in subterranean environments. In both simulation and experiment, ERRT provided efficient exploration and robot-safe paths throughout the exploration mission. The framework was also compared to two other state-of-the-art exploration frameworks and showed superior results in efficiently exploring large-scale environments in time-constrained missions. 

\bibliographystyle{IEEEtran.bst}
\bibliography{mybib}

\begin{thebibliography}{10}
\providecommand{\url}[1]{#1}
\csname url@samestyle\endcsname
\providecommand{\newblock}{\relax}
\providecommand{\bibinfo}[2]{#2}
\providecommand{\BIBentrySTDinterwordspacing}{\spaceskip=0pt\relax}
\providecommand{\BIBentryALTinterwordstretchfactor}{4}
\providecommand{\BIBentryALTinterwordspacing}{\spaceskip=\fontdimen2\font plus
\BIBentryALTinterwordstretchfactor\fontdimen3\font minus \fontdimen4\font\relax}
\providecommand{\BIBforeignlanguage}[2]{{%
\expandafter\ifx\csname l@#1\endcsname\relax
\typeout{** WARNING: IEEEtran.bst: No hyphenation pattern has been}%
\typeout{** loaded for the language `#1'. Using the pattern for}%
\typeout{** the default language instead.}%
\else
\language=\csname l@#1\endcsname
\fi
#2}}
\providecommand{\BIBdecl}{\relax}
\BIBdecl

\bibitem{dang2020autonomous}
T.~Dang, F.~Mascarich, S.~Khattak, H.~Nguyen, H.~Nguyen, S.~Hirsh, R.~Reinhart, C.~Papachristos, and K.~Alexis, ``Autonomous search for underground mine rescue using aerial robots,'' in \emph{2020 IEEE Aerospace Conference}.\hskip 1em plus 0.5em minus 0.4em\relax IEEE, 2020, pp. 1--8.

\bibitem{lindqvist2022compra}
B.~Lindqvist, C.~Kanellakis, S.~S. Mansouri, A.-a. Agha-mohammadi, and G.~Nikolakopoulos, ``Compra: A compact reactive autonomy framework for subterranean mav based search-and-rescue operations,'' \emph{Journal of Intelligent \& Robotic Systems}, vol. 105, no.~3, p.~49, 2022.

\bibitem{petravcek2021large}
P.~Petr{\'a}{\v{c}}ek, V.~Kr{\'a}tk{\`y}, M.~Petrl{\'\i}k, T.~B{\'a}{\v{c}}a, R.~Kratochv{\'\i}l, and M.~Saska, ``Large-scale exploration of cave environments by unmanned aerial vehicles,'' \emph{IEEE Robotics and Automation Letters}, vol.~6, no.~4, pp. 7596--7603, 2021.

\bibitem{kratky2021autonomous}
V.~Kr{\'a}tk{\`y}, P.~Petr{\'a}{\v{c}}ek, T.~B{\'a}{\v{c}}a, and M.~Saska, ``An autonomous unmanned aerial vehicle system for fast exploration of large complex indoor environments,'' \emph{Journal of field robotics}, vol.~38, no.~8, pp. 1036--1058, 2021.

\bibitem{subt}
\BIBentryALTinterwordspacing
DARPA. Subterranean challenge {(SubT)}. [Online]. Available: \url{https://www.subtchallenge.com/}
\BIBentrySTDinterwordspacing

\bibitem{agha2021nebula}
A.~Agha, K.~Otsu, B.~Morrell, D.~D. Fan, R.~Thakker, A.~Santamaria-Navarro, S.-K. Kim, A.~Bouman, X.~Lei, J.~Edlund \emph{et~al.}, ``Nebula: Quest for robotic autonomy in challenging environments; team costar at the darpa subterranean challenge,'' \emph{arXiv preprint arXiv:2103.11470}, 2021.

\bibitem{rouvcek2019darpa}
T.~Rou{\v{c}}ek, M.~Pecka, P.~{\v{C}}{\'\i}{\v{z}}ek, T.~Pet{\v{r}}{\'\i}{\v{c}}ek, J.~Bayer, V.~{\v{S}}alansk{\`y}, D.~He{\v{r}}t, M.~Petrl{\'\i}k, T.~B{\'a}{\v{c}}a, V.~Spurn{\`y} \emph{et~al.}, ``Darpa subterranean challenge: Multi-robotic exploration of underground environments,'' in \emph{International Conference on Modelling and Simulation for Autonomous Systesm}.\hskip 1em plus 0.5em minus 0.4em\relax Springer, Cham, 2019, pp. 274--290.

\bibitem{tranzatto2022cerberus}
M.~Tranzatto, T.~Miki, M.~Dharmadhikari, L.~Bernreiter, M.~Kulkarni, F.~Mascarich, O.~Andersson, S.~Khattak, M.~Hutter, R.~Siegwart \emph{et~al.}, ``Cerberus in the darpa subterranean challenge,'' \emph{Science Robotics}, vol.~7, no.~66, p. eabp9742, 2022.

\bibitem{yamauchi1997frontier}
B.~Yamauchi, ``A frontier-based approach for autonomous exploration,'' in \emph{Proceedings 1997 IEEE International Symposium on Computational Intelligence in Robotics and Automation CIRA'97.'Towards New Computational Principles for Robotics and Automation'}.\hskip 1em plus 0.5em minus 0.4em\relax IEEE, 1997, pp. 146--151.

\bibitem{lluvia2021active}
I.~Lluvia, E.~Lazkano, and A.~Ansuategi, ``Active mapping and robot exploration: A survey,'' \emph{Sensors}, vol.~21, no.~7, p. 2445, 2021.

\bibitem{niroui2019deep}
F.~Niroui, K.~Zhang, Z.~Kashino, and G.~Nejat, ``Deep reinforcement learning robot for search and rescue applications: Exploration in unknown cluttered environments,'' \emph{IEEE Robotics and Automation Letters}, vol.~4, no.~2, pp. 610--617, 2019.

\bibitem{duchoe2014path}
F.~Ducho{\.E}, A.~Babineca, M.~Kajana, P.~Be{\.E}oa, M.~Floreka, T.~Ficoa, and L.~Juri{\v{s}}icaa, ``Path planning with modified a star algorithm for a mobile robot,'' \emph{Procedia Engineering}, vol.~96, pp. 59--69, 2014.

\bibitem{patel2023ref}
A.~Patel, B.~Lindqvist, C.~Kanellakis, A.-a. Agha-mohammadi, and G.~Nikolakopoulos, ``Ref: A rapid exploration framework for deploying autonomous mavs in unknown environments,'' \emph{Journal of Intelligent \& Robotic Systems}, vol. 108, no.~3, p.~35, 2023.

\bibitem{karlsson2023d+}
S.~Karlsson, A.~Koval, C.~Kanellakis, and G.~Nikolakopoulos, ``D$^*_+$: A risk aware platform agnostic heterogeneous path planner,'' \emph{Expert systems with applications}, vol. 215, p. 119408, 2023.

\bibitem{zhou2021fuel}
B.~Zhou, Y.~Zhang, X.~Chen, and S.~Shen, ``Fuel: Fast uav exploration using incremental frontier structure and hierarchical planning,'' \emph{IEEE Robotics and Automation Letters}, vol.~6, no.~2, pp. 779--786, 2021.

\bibitem{zhou2019robust}
B.~Zhou, F.~Gao, L.~Wang, C.~Liu, and S.~Shen, ``Robust and efficient quadrotor trajectory generation for fast autonomous flight,'' \emph{IEEE Robotics and Automation Letters}, vol.~4, no.~4, pp. 3529--3536, 2019.

\bibitem{liu2022efficient}
J.~Liu, Y.~Lv, Y.~Yuan, W.~Chi, G.~Chen, and L.~Sun, ``An efficient robot exploration method based on heuristics biased sampling,'' \emph{IEEE Transactions on Industrial Electronics}, 2022.

\bibitem{pito1999solution}
R.~Pito, ``A solution to the next best view problem for automated surface acquisition,'' \emph{IEEE Transactions on pattern analysis and machine intelligence}, vol.~21, no.~10, pp. 1016--1030, 1999.

\bibitem{low2006efficient}
K.-L. Low and A.~Lastra, ``Efficient constraint evaluation algorithms for hierarchical next-best-view planning,'' in \emph{Third International Symposium on 3D Data Processing, Visualization, and Transmission (3DPVT'06)}.\hskip 1em plus 0.5em minus 0.4em\relax IEEE, 2006, pp. 830--837.

\bibitem{kuffner2000rrt}
J.~J. Kuffner and S.~M. LaValle, ``Rrt-connect: An efficient approach to single-query path planning,'' in \emph{Proceedings 2000 ICRA. Millennium Conference. IEEE International Conference on Robotics and Automation. Symposia Proceedings (Cat. No. 00CH37065)}, vol.~2.\hskip 1em plus 0.5em minus 0.4em\relax IEEE, 2000, pp. 995--1001.

\bibitem{bircher2016receding}
A.~Bircher, M.~Kamel, K.~Alexis, H.~Oleynikova, and R.~Siegwart, ``Receding horizon" next-best-view" planner for 3d exploration,'' in \emph{2016 IEEE international conference on robotics and automation (ICRA)}.\hskip 1em plus 0.5em minus 0.4em\relax IEEE, 2016, pp. 1462--1468.

\bibitem{vasquez2018tree}
J.~I. Vasquez-Gomez, L.~E. Sucar, R.~Murrieta-Cid, and J.-C. Herrera-Lozada, ``Tree-based search of the next best view/state for three-dimensional object reconstruction,'' \emph{International Journal of Advanced Robotic Systems}, vol.~15, no.~1, p. 1729881418754575, 2018.

\bibitem{schmid2020efficient}
L.~Schmid, M.~Pantic, R.~Khanna, L.~Ott, R.~Siegwart, and J.~Nieto, ``An efficient sampling-based method for online informative path planning in unknown environments,'' \emph{IEEE Robotics and Automation Letters}, vol.~5, no.~2, pp. 1500--1507, 2020.

\bibitem{dharmadhikari2020motion}
M.~Dharmadhikari, T.~Dang, L.~Solanka, J.~Loje, H.~Nguyen, N.~Khedekar, and K.~Alexis, ``Motion primitives-based path planning for fast and agile exploration using aerial robots,'' in \emph{2020 IEEE International Conference on Robotics and Automation (ICRA)}.\hskip 1em plus 0.5em minus 0.4em\relax IEEE, 2020, pp. 179--185.

\bibitem{dang2019graph}
T.~Dang, F.~Mascarich, S.~Khattak, C.~Papachristos, and K.~Alexis, ``Graph-based path planning for autonomous robotic exploration in subterranean environments,'' in \emph{2019 IEEE/RSJ International Conference on Intelligent Robots and Systems (IROS)}.\hskip 1em plus 0.5em minus 0.4em\relax IEEE, 2019, pp. 3105--3112.

\bibitem{dang2020graph}
T.~Dang, M.~Tranzatto, S.~Khattak, F.~Mascarich, K.~Alexis, and M.~Hutter, ``Graph-based subterranean exploration path planning using aerial and legged robots,'' \emph{Journal of Field Robotics}, vol.~37, no.~8, pp. 1363--1388, 2020.

\bibitem{hollinger2014sampling}
G.~A. Hollinger and G.~S. Sukhatme, ``Sampling-based robotic information gathering algorithms,'' \emph{The International Journal of Robotics Research}, vol.~33, no.~9, pp. 1271--1287, 2014.

\bibitem{viseras2017online}
A.~Viseras, D.~Shutin, and L.~Merino, ``Online information gathering using sampling-based planners and gps: An information theoretic approach,'' in \emph{2017 IEEE/RSJ International Conference on Intelligent Robots and Systems (IROS)}.\hskip 1em plus 0.5em minus 0.4em\relax IEEE, 2017, pp. 123--130.

\bibitem{lindqvist2021exploration}
B.~Lindqvist, A.-A. Agha-Mohammadi, and G.~Nikolakopoulos, ``Exploration-rrt: A multi-objective path planning and exploration framework for unknown and unstructured environments,'' in \emph{2021 IEEE/RSJ International Conference on Intelligent Robots and Systems (IROS)}.\hskip 1em plus 0.5em minus 0.4em\relax IEEE, 2021, pp. 3429--3435.

\bibitem{hornung2013octomap}
A.~Hornung, K.~M. Wurm, M.~Bennewitz, C.~Stachniss, and W.~Burgard, ``Octomap: An efficient probabilistic 3d mapping framework based on octrees,'' \emph{Autonomous robots}, vol.~34, no.~3, pp. 189--206, 2013.

\bibitem{musil2022spheremap}
T.~Musil, M.~Petrl{\'\i}k, and M.~Saska, ``Spheremap: Dynamic multi-layer graph structure for rapid safety-aware uav planning,'' \emph{IEEE Robotics and Automation Letters}, vol.~7, no.~4, pp. 11\,007--11\,014, 2022.

\bibitem{cover2013sparse}
H.~Cover, S.~Choudhury, S.~Scherer, and S.~Singh, ``Sparse tangential network (spartan): Motion planning for micro aerial vehicles,'' in \emph{2013 IEEE International Conference on Robotics and Automation}.\hskip 1em plus 0.5em minus 0.4em\relax IEEE, 2013, pp. 2820--2825.

\bibitem{duberg2020ufomap}
D.~Duberg and P.~Jensfelt, ``Ufomap: An efficient probabilistic 3d mapping framework that embraces the unknown,'' \emph{IEEE Robotics and Automation Letters}, vol.~5, no.~4, pp. 6411--6418, 2020.

\bibitem{quigley2009ros}
M.~Quigley, K.~Conley, B.~Gerkey, J.~Faust, T.~Foote, J.~Leibs, R.~Wheeler, and A.~Y. Ng, ``{ROS}: an open-source robot operating system,'' in \emph{ICRA workshop on open source software}, vol.~3.\hskip 1em plus 0.5em minus 0.4em\relax Kobe, Japan, 2009, p.~5.

\bibitem{sopasakis2020open}
P.~Sopasakis, E.~Fresk, and P.~Patrinos, ``Open: Code generation for embedded nonconvex optimization,'' \emph{International Federation of Automatic Control}, 2020.

\bibitem{wang2020neural}
J.~Wang, W.~Chi, C.~Li, C.~Wang, and M.~Q.-H. Meng, ``Neural rrt*: Learning-based optimal path planning,'' \emph{IEEE Transactions on Automation Science and Engineering}, vol.~17, no.~4, pp. 1748--1758, 2020.

\bibitem{otte2016rrtx}
M.~Otte and E.~Frazzoli, ``Rrtx: Asymptotically optimal single-query sampling-based motion planning with quick replanning,'' \emph{The International Journal of Robotics Research}, vol.~35, no.~7, pp. 797--822, 2016.

\bibitem{qi2020mod}
J.~Qi, H.~Yang, and H.~Sun, ``Mod-rrt*: A sampling-based algorithm for robot path planning in dynamic environment,'' \emph{IEEE Transactions on Industrial Electronics}, vol.~68, no.~8, pp. 7244--7251, 2020.

\bibitem{noreen2016optimal}
I.~Noreen, A.~Khan, and Z.~Habib, ``Optimal path planning using rrt* based approaches: a survey and future directions,'' \emph{International Journal of Advanced Computer Science and Applications}, vol.~7, no.~11, 2016.

\bibitem{sathya2018embedded}
A.~Sathya, P.~Sopasakis, R.~Van~Parys, A.~Themelis, G.~Pipeleers, and P.~Patrinos, ``Embedded nonlinear model predictive control for obstacle avoidance using panoc,'' in \emph{2018 European Control Conference (ECC)}.\hskip 1em plus 0.5em minus 0.4em\relax IEEE, 2018, pp. 1523--1528.

\bibitem{small2019aerial}
E.~Small, P.~Sopasakis, E.~Fresk, P.~Patrinos, and G.~Nikolakopoulos, ``Aerial navigation in obstructed environments with embedded nonlinear model predictive control,'' in \emph{2019 18th European Control Conference (ECC)}.\hskip 1em plus 0.5em minus 0.4em\relax IEEE, 2019, pp. 3556--3563.

\bibitem{lindqvist2020dynamic}
B.~Lindqvist, S.~S. Mansouri, A.-a. Agha-mohammadi, and G.~Nikolakopoulos, ``Nonlinear mpc for collision avoidance and control of uavs with dynamic obstacles,'' \emph{IEEE Robotics and Automation Letters}, vol.~5, no.~4, pp. 6001--6008, 2020.

\bibitem{stella2017simple}
L.~Stella, A.~Themelis, P.~Sopasakis, and P.~Patrinos, ``A simple and efficient algorithm for nonlinear model predictive control,'' in \emph{2017 IEEE 56th Annual Conference on Decision and Control (CDC)}.\hskip 1em plus 0.5em minus 0.4em\relax IEEE, 2017, pp. 1939--1944.

\bibitem{furrer2016rotors}
F.~Furrer, M.~Burri, M.~Achtelik, and R.~Siegwart, ``Rotors—a modular gazebo mav simulator framework,'' \emph{Robot Operating System (ROS) The Complete Reference (Volume 1)}, pp. 595--625, 2016.

\bibitem{meier2011pixhawk}
L.~Meier, P.~Tanskanen, F.~Fraundorfer, and M.~Pollefeys, ``Pixhawk: A system for autonomous flight using onboard computer vision,'' in \emph{2011 IEEE International Conference on Robotics and Automation}.\hskip 1em plus 0.5em minus 0.4em\relax IEEE, 2011, pp. 2992--2997.

\bibitem{meier2015px4}
L.~Meier, D.~Honegger, and M.~Pollefeys, ``Px4: A node-based multithreaded open source robotics framework for deeply embedded platforms,'' in \emph{2015 IEEE international conference on robotics and automation (ICRA)}.\hskip 1em plus 0.5em minus 0.4em\relax IEEE, 2015, pp. 6235--6240.

\bibitem{lindqvist2022adaptive}
B.~Lindqvist, J.~Haluska, C.~Kanellakis, and G.~Nikolakopoulos, ``An adaptive 3d artificial potential field for fail-safe uav navigation,'' in \emph{2022 30th Mediterranean Conference on Control and Automation (MED)}.\hskip 1em plus 0.5em minus 0.4em\relax IEEE, 2022, pp. 362--367.

\bibitem{shan2020lio}
T.~Shan, B.~Englot, D.~Meyers, W.~Wang, C.~Ratti, and D.~Rus, ``{LIO-SAM: Tightly-coupled Lidar Inertial Odometry via Smoothing and Mapping},'' in \emph{IEEE/RSJ International Conference on Intelligent Robots and Systems (IROS)}, 2020.

\bibitem{carrillo2015autonomous}
H.~Carrillo, P.~Dames, V.~Kumar, and J.~A. Castellanos, ``Autonomous robotic exploration using occupancy grid maps and graph slam based on shannon and r{\'e}nyi entropy,'' in \emph{2015 IEEE international conference on robotics and automation (ICRA)}.\hskip 1em plus 0.5em minus 0.4em\relax IEEE, 2015, pp. 487--494.

\bibitem{carrillo2018autonomous}
------, ``Autonomous robotic exploration using a utility function based on r{\'e}nyi’s general theory of entropy,'' \emph{Autonomous Robots}, vol.~42, no.~2, pp. 235--256, 2018.

\bibitem{papachristos2019localization}
C.~Papachristos, F.~Mascarich, S.~Khattak, T.~Dang, and K.~Alexis, ``Localization uncertainty-aware autonomous exploration and mapping with aerial robots using receding horizon path-planning,'' \emph{Autonomous Robots}, vol.~43, pp. 2131--2161, 2019.

\end{thebibliography}

\begin{IEEEbiography}
    [{\includegraphics[width=1in,height=1.25in,clip,keepaspectratio]{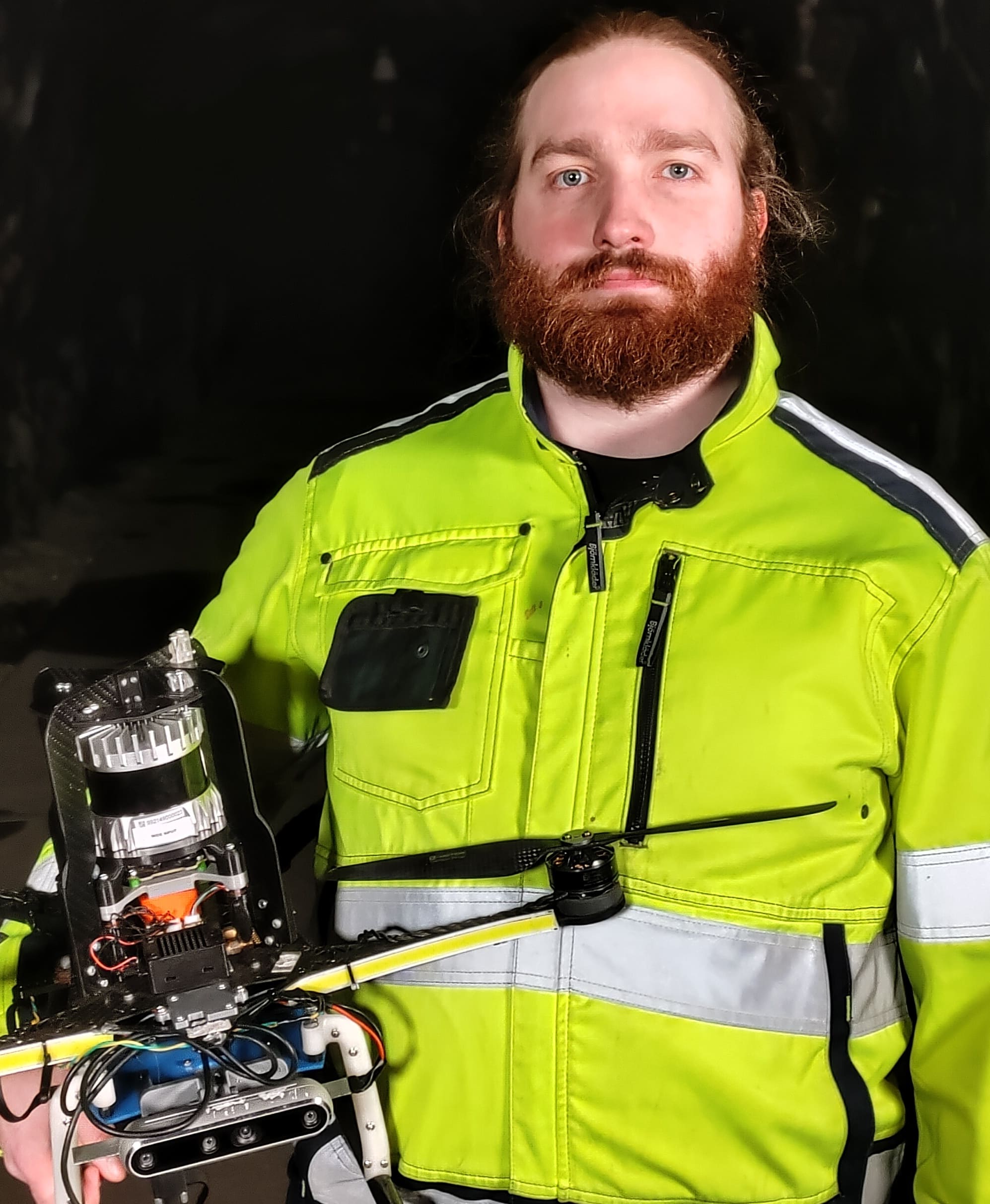}}]{Björn Lindqvist}
received his PhD in 2023 at the Robotics and AI Group at the Department of Computer Science, Electrical and Space Engineering, Luleå University of Technology, Sweden, from his work on robotic reactive navigation methods. He received his Master's Degree in Space Engineering with a specialisation Aerospace Engineering from Luleå University of Technology, Sweden, in 2019. Björn's research has so far been focused on collision avoidance and path planning for single and multi-agent Unmanned Aerial Vehicle systems, as well as field applications of such technologies. He has worked as part of the JPL-NASA led Team CoSTAR in the DARPA Sub-T Challenge on subterranean UAV exploration applications, specifically in the search-and-rescue context. He has also worked in multiple national and EU projects related to the deployment of autonomous UAV in mining environments.
\end{IEEEbiography}

\begin{IEEEbiography}
    [{\includegraphics[width=1in,height=1.25in,clip,keepaspectratio]{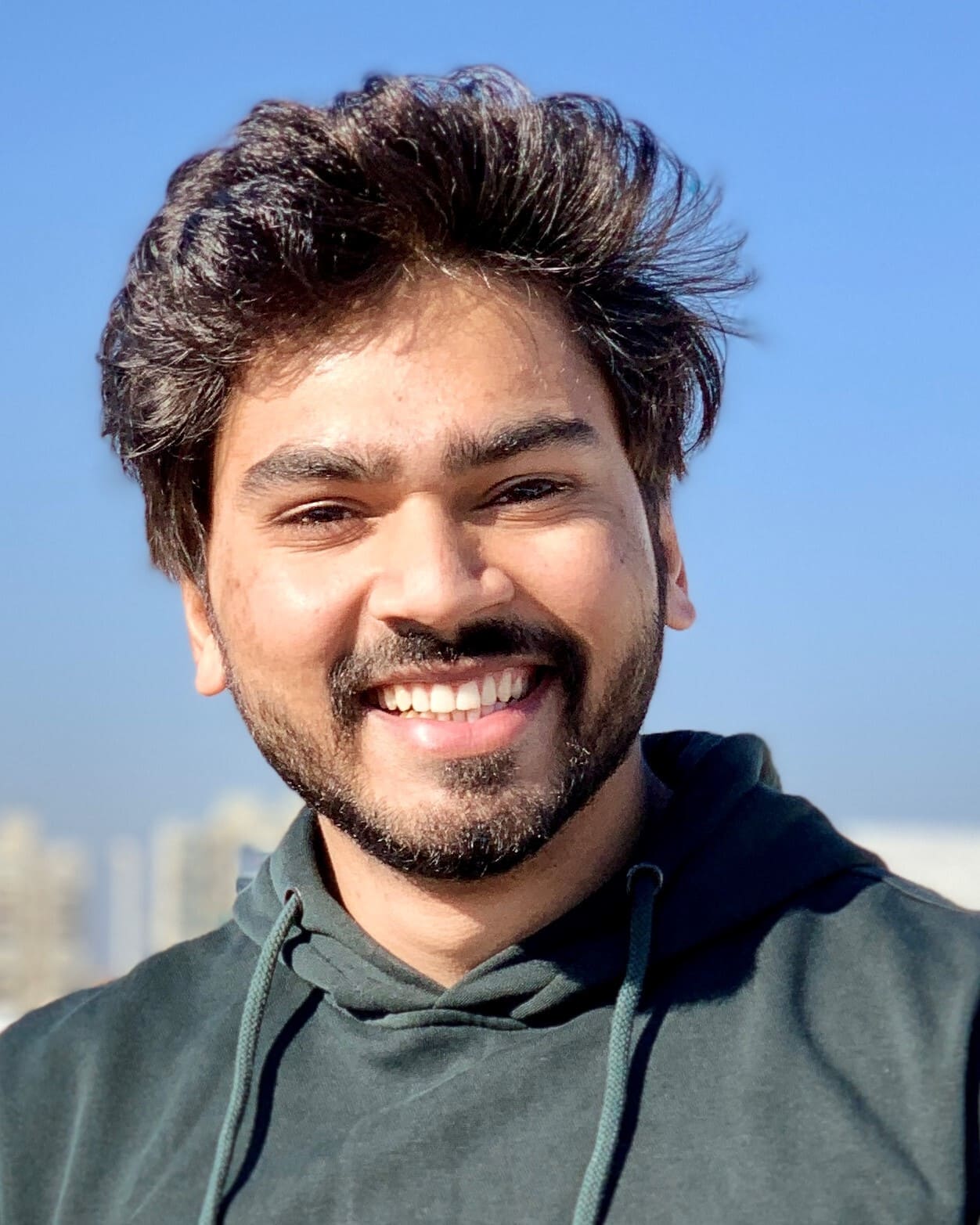}}]{Akash Patel}
 is currently pursuing his PhD at the Robotics and AI Group of the Luleå University of Technology, Sweden. His current research direction is focused on developing exploration and mapping algorithms to enable autonomous exploration of caves, lava tubes and voids of the planetary bodies. Akash received his master’s degree in Space Science and Technology from Luleå University of Technology, Sweden and bachelor’s degree in Aerospace Engineering from UPES, India.
\end{IEEEbiography}

\begin{IEEEbiography}
    [{\includegraphics[width=1in,height=1.25in,clip,keepaspectratio]{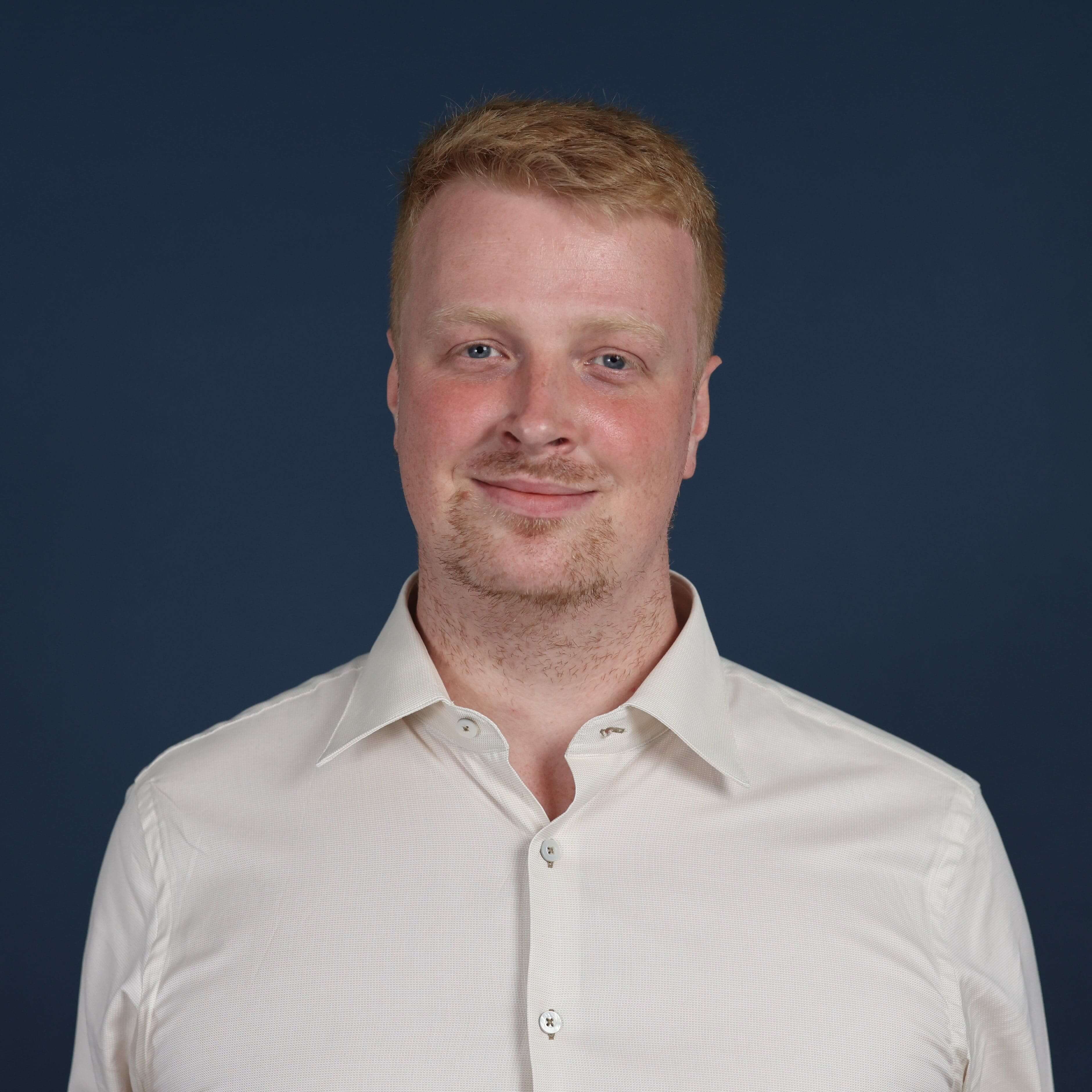}}]{Kalle Löfgren}
 received his Master's Degree in Computer Science Engineering from Luleå University of Technology, Sweden, in 2022 with a specialisation in Industrial Computer Systems. He did his Master Thesis at the Robotics and Artificial Intelligence Group at Luleå University of Technology, focusing on real-time path planning and robot exploration, and code optimization. 
\end{IEEEbiography}

\begin{IEEEbiography}
    [{\includegraphics[width=1in,height=1.25in,clip,keepaspectratio]{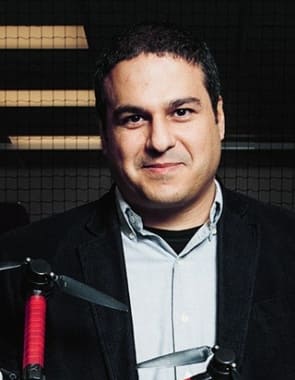}}]{Prof. George Nikolakopoulos}
works as the Chair Professor in Robotics and Artificial Intelligence, at the Department of Computer Science, Electrical and Space Engineering, Luleå University of Technology (LTU), Sweden. He was also affiliated with the NASA Jet Propulsion Laboratory for conducting collaborative research on Aerial Planetary Exploration and participated in the DARPA Grand challenge on Sub-T exploration with the CoSTAR team of JPL-NASA. He is a Director of euRobotics, and is a member of the Scientific Council of ARTEMIS and the IFAC TC on Robotics. He established the Digital Innovation Hub on Applied AI at LTU and also represents Sweden in the Technical Expert Group on Robotics and AI in the project MIRAI 2.0 promoting collaboration between Sweden and Japan. His main research interests are in the areas of: Field Robotics, Space Autonomy, UAVs, Automatic Control Applications, Networked Embedded Controlled Systems, Wireless Sensor and Actuator Networks, Cyber Physical Systems, and Adaptive Control.
\end{IEEEbiography}

\end{document}